\documentclass[acmsmall]{acmart}
\usepackage{algorithm}
\usepackage{algpseudocode}
\usepackage{amsmath}
\usepackage{tikz}
\usepackage{array}
\usepackage{balance}
\usepackage{enumitem}
\usepackage{fancyvrb}   
\usepackage{fbox}
\usepackage{graphicx}  
\usepackage{listings}
\usepackage{longtable}
\usepackage{multirow}
\usepackage{booktabs}
\usepackage{tabularx}
\usepackage[most]{tcolorbox}
\usepackage{textcomp}
\usepackage{varwidth}   
\usepackage{wrapfig}
\usepackage[dvipsnames]{xcolor}
\usepackage{colortbl}
\usetikzlibrary{shadows}

\newcommand{\mypar}[1]{\vspace{0.15cm} \noindent {\textbf{#1.}\/}}

\definecolor{DarkOrange}{rgb}{0.8,0.4,0}
\definecolor{DarkRed}{rgb}{0.6,0,0}
\definecolor{ForestGreen}{rgb}{0.13,0.55,0.13}
\definecolor{Purple}{rgb}{0.70,0.0,0.70}
\definecolor{linegray}{rgb}{0.93,0.93,0.93}
\definecolor{lineblue}{rgb}{0.9,0.9,0.99}
\definecolor{linegreen}{rgb}{0.91,0.95,0.91}
\definecolor{linegreen2}{rgb}{0.98,1.0,0.98}
\newcommand{\tinycolorbox}[2]{\tikz[baseline=(a.base),inner sep=0pt]\node[fill=#1](a){#2};}

\newcommand{\anomaly}[1]{\textbf{\textcolor{DarkOrange}{#1}}}
\newcommand{\rootcause}[1]{\textbf{\textcolor{DarkRed}{#1}}}

\newcommand{\circled}[1]{\raisebox{1pt}{\tikz[baseline=(char.base)]{%
  \node[shape=circle, fill=black, text=white, inner sep=1pt, font=\scriptsize\bfseries] (char) {#1};}}}

\def\toolname{\mbox{\textsc{EventADL}}}

\def\mxpEvent{AVA}
\def\zrhEvent{OUT}
\def\aioEvent{AIO}

\def\aws{UKW}
\def\iam{Identity}

\lstdefinelanguage{json}{
  morestring=[b]",
  morecomment=[l]{//},
  moredelim=[l][\color{black}]{:},
  moredelim=[s][\color{red}]{[}{]},
  stringstyle=\color{teal},
  keywordstyle=\color{black},
  commentstyle=\color{gray},
  showstringspaces=false,
}
\newcolumntype{L}{>{\raggedright\arraybackslash}X}

\setcopyright{cc}
\setcctype{by-nc-nd}
\acmDOI{10.1145/3808186}
\acmYear{2026}
\acmJournal{PACMSE}
\acmVolume{3}
\acmNumber{FSE}
\acmArticle{FSE179}
\acmMonth{7}
\acmSubmissionID{fse26mainb-p1676-p}
\received{2025-09-11}
\received[accepted]{2026-03-24}

\begin{document}

\title{\toolname{}: Open-Box Anomaly Detection and Localization Framework for Events in Cloud-Based Service Systems}

\author{Luan Pham}
\orcid{0000-0001-7243-3225}
\affiliation{%
  \institution{RMIT University}
  \country{Australia}
}
\email{luan.pham@rmit.edu.au}

\author{Victor Nicolet}
\orcid{0000-0002-3743-7498}
\affiliation{%
  \institution{Amazon Web Services}
  \country{United States}
}
\email{victornl@amazon.com}

\author{Joey Dodds}
\orcid{0009-0004-1534-6968}
\affiliation{%
  \institution{Amazon Web Services}
  \country{United States}
}
\email{jldodds@amazon.com}

\author{Hui Guan}
\orcid{0000-0001-9128-2231}
\affiliation{%
  \institution{Amazon Web Services}
  \country{United States}
}
\email{huiguan@amazon.com}

\author{Daniel Kroening}
\orcid{0000-0002-6681-5283}
\affiliation{%
  \institution{Amazon Web Services}
  \country{United States}
}
\email{dkr@amazon.com}

\begin{abstract}
Anomaly detection and localization (ADL) is critical for maintaining reliability and availability in cloud systems. Recent ADL developments focus on metric and log data, leaving event data unexplored. To address this gap, we propose \toolname{}, the first \textit{open-box} event-based ADL framework for cloud-based service systems. To motivate the design of our framework, we conduct a systematic analysis on 520 real-world incidents, and provide insights into how anomalies and their root causes manifest through event data. \toolname{} has three phases: offline training, online anomaly detection, and root cause localization. During the training phase, \toolname{} first learns \emph{Event Semantic Patterns (ESPs)}, which capture normal interactions between system entities using historical event data, and then learns \emph{Event Frequency Patterns (EFPs)}, which capture the normal frequency of known ESPs. In the online anomaly detection phase, any data in the event stream that deviates significantly from either pattern is identified as anomalous. For localization, \toolname{} constructs an \textit{Intervention Graph} that models the relationships between recent system interactions and the detected anomalies for automatic root cause localization. The framework is designed to operate efficiently with unlabeled data and to produce interpretable anomalies with their corresponding root causes. Our evaluation on three real cloud service systems and two real-world incidents demonstrates that \toolname{} outperforms existing methods, achieving F1-scores of at least 90\% for anomaly detection and 100\% top-3 accuracy in root cause localization.
\end{abstract}



\maketitle

\section{Introduction}

Cloud-based service systems generate large volumes of structured event data that record who performed what operation on which resources and when. These events are captured by monitoring systems to track operations such as API calls, configuration changes, and resource updates. Leading cloud providers offer event monitoring services to support observability and auditing, including AWS CloudTrail~\cite{aws_cloudtrail_events}, Azure Event Hub~\cite{azure_monitor_eventhub_ingestion}, Google Cloud Audit~\cite{google_cloud_audit_logs}, and Alibaba ActionTrail~\cite{alibaba-actiontrail}.

A fundamental challenge faced by cloud operators is \textbf{timely anomaly detection and localization (ADL)} from event streams. Anomaly detection identifies unexpected behaviors in the system, while root cause localization (RCL) pinpoints the root causes of those anomalies. Notably, the root cause may not be anomalous on its own. For instance, a resource deletion may seem benign in isolation but could trigger catastrophic failures in downstream services depending on the deleted resource. Our study shows that detecting and diagnosing such incidents often involves multiple teams and significant effort to analyze observable event data. Given the operational scale and high stakes, automated and interpretable ADL for event data is a critical capability in modern cloud~\cite{cheng2023ai}.

\begin{wrapfigure}{r}{0.38\textwidth}
\vspace{-10pt}
\centering
\begin{lstlisting}[language=json, basicstyle=\ttfamily\scriptsize, backgroundcolor=\color{gray!10}, frame=single, columns=fullflexible, keepspaces=true, xrightmargin=5pt]
{ "actor.user.name": "merlinary",
  "api.operation": "UpdateInstances",
  "api.request.data": {
    "force": true
  },
  "resources": [
    { "uid": "prod-04242345432" }
  ],
  "cloud.region": "us-east-1",
  "time": "2025-05-19T17:38:32Z",
  "error": null
}
\end{lstlisting}
\vspace{-10pt}
\caption{An event in the OCSF schema~\cite{ocsf}.} \label{fig:event-example}
\vspace{-10pt}
\end{wrapfigure}
\vspace{-3pt}
\mypar{Limitations of Existing Work}
While ADL has been actively studied for metrics~\cite{pham2026graph, gu2024kpiroot, pham2024baro, gu2025adamas, circa, chen2022adsketch} and unstructured logs~\cite{brianlogsurvey2025, le2021neurallog, du2017deeplog, landauer2024critical, meng2019loganomaly, li2020swisslog}, event-based ADL remains underexplored. Metric-based methods~\cite{pham2024baro, circa, pham2026graph} capture frequency-based anomalies but fail to detect pointwise anomalies (i.e., an individual anomalous event). Log-based methods such as DeepLog~\cite{du2017deeplog}, LogAnomaly~\cite{meng2019loganomaly}, and SwissLog~\cite{li2020swisslog} employ deep neural networks (LSTMs and BERT) to capture sequential, quantitative, and temporal patterns in log data. These methods suffer from four key limitations: \circled{1}~they are \textit{closed-box}, offering no interpretability into why an anomaly was flagged; \circled{2}~they perform detection only, without localizing root causes; \circled{3}~they lack adaptive mechanisms for evolving systems, requiring retraining when system behavior changes; and \circled{4}~they ignore structured information readily available in events, such as the interactions between actors and resources (Figure~\ref{fig:event-example}).

Most event-based anomaly detection methods~\cite{zengy2022shadewatcher, amin2019cadence, coskun2022detecting} also rely on deep neural networks and thus cannot offer interpretable decisions. ShadeWatcher~\cite{zengy2022shadewatcher} uses a context-aware embedding model and a GNN to identify anomalous interactions, with anomaly scores derived from deep GNN embeddings. GuardDuty~\cite{coskun2022detecting} employs a VAE trained on historical data to detect anomalies based on reconstruction error in the latent space. These systems provide no explanation for why an anomaly is detected, requiring further manual investigation.
HyGLAD~\cite{hyglad} is a pointwise anomaly detection technique which is unable to detect frequency-based anomalies.

Most existing RCL methods focus on metrics, logs, and traces. Nezha~\cite{yu2023nezha}, MULAN~\cite{zheng2024mulan}, MRCA~\cite{wang2024mrca}, and CORAL~\cite{wang2023incremental} construct causal graphs from metrics, logs, traces and apply centrality or causality analysis for RCL. CONAN~\cite{li2023conan} extracts contrast patterns from attribute-value pairs to diagnose batch failures. TraceContrast~\cite{zhang2024trace} mines contrast sequential patterns from distributed traces. SLIM~\cite{ren2024slim} uses supervised rule-set learning from fault data, and ReACT~\cite{roy2024exploring} leverages LLMs to reason over incident reports. While these methods advance RCL in their respective domains, none are designed for structured event data that encodes explicit actor-operation-resource relationships. Their data representations lack the semantic structure that enables open-box detection of \textit{Event Type}, \textit{Event Value}, and \textit{Event Frequency} anomalies, and interpretable RCL through intervention graphs.

\vspace{-3pt}
\mypar{Empirical Study} Despite the importance of event data, no prior work has systematically analyzed how anomalies and their root causes manifest through events. We address this gap with a novel empirical study of 520 real-world incident reports from production systems (Section~\ref{sec:analysis-real-world}). This analysis provides the empirical foundation that has been missing from prior work, offering guidance for designing effective event-based ADL techniques. Our analysis reveals two key findings. First, event-based anomalies manifest along three dimensions: \textit{Event Type (21\%)}, \textit{Event Value (68\%)}, and \textit{Event Frequency (67\%)}. This distribution indicates that effective detection requires capturing both semantic deviations (unusual types or values) and temporal deviations (abnormal frequencies). Second, root causes stem from either a single intervention (\textit{32\%}) or multiple interventions (\textit{68\%}), such as resource deletions or configuration changes. This pattern suggests that RCL must trace causal chains from observed anomalies back through potentially multiple triggering events.

\vspace{-3pt}
\mypar{Proposed Approach} Guided by these findings, we present \toolname{}, the first \textit{open-box} ADL framework tailored for event data. \toolname{} provides both anomaly detection and RCL capabilities while maintaining \textbf{interpretability} in two ways. First, when anomalies are detected, operators receive not just alerts but also interpretable explanations that detail \textit{why} these anomalies occurred. Second, all components of \toolname{} are interpretable and transparent, enabling operators to validate the results and build trust in automated diagnostics. \toolname{} operates in three phases: offline training, online anomaly detection, and RCL. During offline training, \toolname{} learns human-interpretable patterns: \textit{Event Semantic Patterns (ESPs)} and \textit{Event Frequency Patterns (EFPs)}. ESPs capture normal event field structures (i.e., Event Types and Values), addressing the Event Type and Event Value anomalies identified in our study. EFPs represent the expected frequency patterns of events, targeting Event Frequency anomalies. During the online detection phase, \toolname{} detects anomalies by comparing incoming events against learned patterns. Violations of ESPs indicate pointwise anomalies, while deviations from EFPs signal frequency-based anomalies. For RCL, \toolname{} constructs an \textit{Intervention Graph} capturing interventions that may have triggered the anomalies. The Intervention Graph models causal relationships between actors, operations, impacted resources, and detected anomalies. \toolname{} applies a time-aware random walk over this graph to identify root causes, tracing the causal chains that our analysis found prevalent.



In summary, we present a comprehensive research effort on event-based ADL, an area that remains underexplored. Our work spans problem formulation, systematic analysis, framework design, extensive evaluation, and dataset release to facilitate future research. Our major contributions are:
\begin{itemize}[leftmargin=*, topsep=1pt]
\item We conduct a systematic analysis of 520 real-world incident reports, offering valuable insights into how anomalies and their root causes manifest in event data in cloud-based service systems.
\item We introduce \toolname{}, the first open-box ADL framework for events in cloud-based service systems. \toolname{} detects anomalies in Event Type, Event Value, and Event Frequency in event data and localizes their corresponding root causes. Our framework is designed to operate efficiently, with unlabeled data, and to provide an interpretable ADL outcome.
\item We extensively evaluate \toolname{} on three benchmark datasets and two real-world incidents. The results show that \toolname{} consistently outperforms existing state-of-the-art methods, achieving F1-scores of at least 90\% for anomaly detection and 100\% top-3 accuracy in RCL.
\end{itemize}

\vspace{-5pt}
\section{Terminology and Problem Statement} \label{sec:background}


\subsection{Terminology} \label{sec:term}

\vspace{-5pt}
\mypar{Event} In cloud systems, an event is a structured record that captures at least four key attributes: the \texttt{actor} (who performed the action), the \texttt{operation} (what action was performed), the \texttt{resources} (what was acted upon), and the \texttt{timestamp} (when the event occurred). Events may include auxiliary attributes such as execution parameters, metadata, or error indicators. Compared to unstructured logs, which require parsing for analysis~\cite{he2017drain, brianlogsurvey2025}, events have a known schema. Events are standardized in both open-source~\cite{opentelemetry_events_2025, ocsf} and commercial~\cite{aws_cloudtrail_events, azure_monitor_eventhub_ingestion, google_cloud_audit_logs} platforms. Figure~\ref{fig:event-example} shows an event following the OCSF schema~\cite{ocsf}. \textit{Event Type} refers to the categorical field that identifies the kind of interaction being recorded (e.g., the \texttt{api.operation} field in Figure~\ref{fig:event-example}, or the \texttt{eventName} field in AWS CloudTrail~\cite{aws_cloudtrail_events}). An Event Type anomaly occurs when an unseen or unexpected interaction type is observed. \textit{Event Value} refers to the attribute values within an event, such as actor identifiers, resource identifiers, or request parameters. An Event Value anomaly occurs when an unusual combination of values appears (e.g., a development user accessing a production database, or an operation performed in an unexpected region). In Figure~\ref{fig:event-example}, the Event Type is \texttt{UpdateInstances}, while Event Values include the actor \texttt{merlinary}, resource \texttt{prod-04242345432}, and region \texttt{us-east-1}.

\vspace{-5pt}
\mypar{Anomaly} We view as anomalous any \textit{significant deviation from expected system behavior}. Anomalies manifest in event data through four types: (1) \textit{Event Type}, (2) \textit{Event Value}, (3) \textit{Event Frequency}, and (4) \textit{Event Order}. Although prior studies~\cite{zengy2022shadewatcher, event_ijcai16_ape, van2004workflow} have examined subsets of these anomalies, none have systematically defined or comprehensively investigated all four. In Section~\ref{sec:analysis-real-world}, we present an analysis of 520 incident reports to assess the prevalence and significance of these anomaly types.

\vspace{-5pt}
\mypar{Root Cause of Anomalies}
The root cause of an anomaly refers to the initial event(s) that trigger abnormal behavior. These are typically \textit{intervention events}, such as configuration changes or code deployments. Notably, a root cause may not be an anomaly itself in isolation. For instance, a user deleting a resource may be consistent with past behavior. However, if this deletion disrupts downstream services, it is causally linked to subsequent anomalies.

\vspace{-5pt}
\subsection{Problem Formulation} 
\label{sec:prob-form}

Let $\mathcal{E} = \langle e_1, e_2, \dots\rangle$ denote a stream of events emitted by a cloud-based service system. Each event $e_i$ is a tuple \( e_i = (\Lambda_i,\, \Omega_i,\, \mathcal{R}_i,\, \tau_i),\) where $\Lambda_i$~is the \texttt{actor}, $\Omega_i$~is the \texttt{operation}, $\mathcal{R}_i$~is the affected \texttt{resources}, and $\tau_i$~is the \texttt{timestamp}. At timestamp $t$, the system observes a window of recent events denoted by \( \mathcal{W}_t = \{ e_i \in \mathcal{E} \mid \tau_i \in [t - \Delta,\, t] \},\) where $\Delta$ is the window size. \textbf{Anomaly detection} determines whether $\mathcal{W}_t$ contains any anomalous behavior. We define a binary indicator $y \in \{0, 1\}$, where $y = 1$ indicates the presence of anomalies in $\mathcal{W}_t$, and $y = 0$ indicates their absence. If any anomalous event is detected in \(\mathcal{W}_t\), then the entire window is considered anomalous. If $y = 1$, the system triggers \textbf{RCL} to identify the root cause(s) using a possibly extended window~$\mathcal{W}_{t}'$. Formally, the task is to identify a subset of events $\mathcal{C}_{t} \subseteq \mathcal{W}_{t}'$ that constitutes the root cause. The final output is the tuple $(y,\, \mathcal{C}_{t})$, where $y$ indicates the presence of anomalies and $\mathcal{C}_{t}$ identifies the root cause(s).

\begin{table*}
\caption{How do \anomaly{anomalies} and their \rootcause{root cause} manifest through event data in real-world incidents?} \label{tab:incidents-main}
\vspace{-10pt} 
\setlength\tabcolsep{4pt}
\scriptsize
\begin{tabular}{p{0.02\textwidth} p{0.13\textwidth} p{0.40\textwidth} p{0.35\textwidth}}
\toprule
\textbf{ID} & \textbf{Title} & \textbf{Anomaly Symptoms} & \textbf{Root Cause Event} \\
\midrule
\mxpEvent & The Availability of Service X drops in Region A. & The events associated with session token creation in Region A \anomaly{suddenly disappeared}. There was a significant increase in API call volume (\(\approx\)14x normal) due to client retries, with \anomaly{almost all of them (99\%) resulting in errors}. No successful token creation was recorded. 
& The root cause event was \rootcause{the deactivation of an access key used by Service X} for encryption and signing keys. This deactivation occurred during a credentials cleanup when the credential management tool incorrectly identified used keys. 
\\ \midrule
\zrhEvent & Sign-in outage in Region B. &  The anomalies were observed as \anomaly{authentication errors} from AuthService with error messages "Session validation didn't return a principal". The frequency of successful authentication events suddenly stopped for Sign-in service in Region B. & The root cause event was \rootcause{the deletion of an \iam{} role} used by Sign-in service to call Service X APIs. This deletion event occurred as part of an infrastructure stack update triggered by the infrastructure pipeline deployment. \\ \midrule
\aioEvent & AIOps customers encountered 4XX errors for 3 days while creating AQuery observations. & X events show \anomaly{a sudden spike of 4XX error responses for API calls}. The errors occurred when users create investigations or observations using AQueryResult events containing newline characters. This resulted in a \anomaly{96\% failure rate in Region A} and \anomaly{100\% failure rate in Region B} for Y-related events.
& The root cause event was \rootcause{the deployment of a code change that introduced an incorrect regex validation pattern} (\verb|^[\S\s]+$| and later \verb|^.*$|) for API input fields, which rejected valid XXLI queries containing newline characters. This change was to address a security requirement. \\ 
\bottomrule
\end{tabular}

\vspace{-10pt} 
\end{table*}

\vspace{-5pt}
\section{Analysis of Real-world Incidents} \label{sec:analysis-real-world}

To guide the design of our framework, we conducted a systematic empirical study of incident reports from a large service provider, You-Know-Where (\aws{}).  Each incident report typically contains a high-level summary, detailed information on anomaly symptom(s), mitigation action(s), diagnosed root cause(s), and the incident impact (e.g., the number of affected users, the incident duration). Focusing on how engineers investigate anomalies and use event data during incident diagnosis helps us understand how anomalies and their root causes manifest in production environments.

\begin{figure*}[t]
\centering
\includegraphics[width=0.93\linewidth]{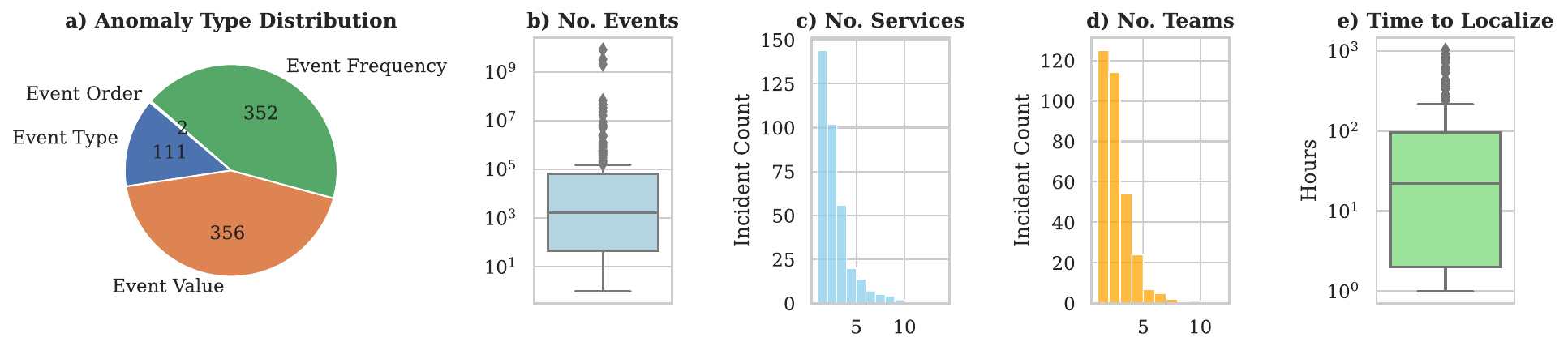}
\vspace{-10pt}
\caption{Insights from real-world incidents. (a) Distribution of anomaly types. (b) Number of events analyzed. (c)~Number of services involved. (d) Number of teams involved. (e)~Root cause localization time.} \label{fig:real-world-insights}
\vspace{-10pt}
\end{figure*}

To systematically select the relevant incident reports, we first queried the incident database using the keywords \texttt{"Service X"} and \texttt{"X Data"}, where X is \aws{}'s event collection service. The search yielded 1{,}634 incidents mentioning \texttt{"Service X"} and 2{,}928 mentioning \texttt{"X Data"} among all incidents since 2013, indicating that event data is a standard part of operational diagnostics. Focusing on recent operational contexts, we restricted our analysis to incidents from the past year 
(June 2024 -- 2025). This search yielded  520 incident reports in total, of which  202 mention \texttt{"Service X"} and  354 mention \texttt{"X Data"}. Our analysis aims to answer two key questions: \textbf{(Q1)}~How do anomalies manifest in event data? and \textbf{(Q2)} How do root causes leave traces in event data?

\vspace{-5pt}
\subsection{Findings} \label{sec:findings}

Figure~\ref{fig:real-world-insights} summarizes our quantitative analysis, and Table~\ref{tab:incidents-main} provides representative case studies.

\vspace{-5pt}
\mypar{Anomaly Symptoms in Event Data (Q1)} We categorized anomaly symptoms using the taxonomy introduced in Section~\ref{sec:term}. We observe that: (1) among the four anomaly types, \textit{Event Type}, \textit{Event Value}, and \textit{Event Frequency} dominate the distribution with only 2 incidents exhibiting \textit{Event Order} anomalies and (2) most incidents exhibit more than one anomaly type. Specifically, 111 incidents (21\%) involve abnormal Event Type; 356 incidents (68\%) involve abnormal Event Value; and 352 incidents (67\%) exhibit anomalies in Event Frequency. Event Value anomalies commonly manifest as unusual actor-resource relationships, such as unauthorized actors accessing protected resources, actors operating in unexpected regions, or actors performing operations on resources outside their scope. The two incidents with abnormal Event Order are caused by secret rotation procedures (e.g., deleting the old secret before the new one becomes available). Most incidents involve multiple anomaly types (72\%; 375 incidents), and 145 incidents exhibit a single anomaly type.

The most frequent combination is Event Value and Event Frequency (29\%). Note that we report the symptoms observed in incident reports; the actual data may sometimes contain more information. For example, we found that the incidents with Event Order anomalies also exhibit anomalies in Event Frequency (e.g., a spike in error events caused by the deleted key). These findings suggest that an anomaly detection solution should monitor for anomalies in Event Type, Event Value and Event Frequency. This would allow the detection of anomalies for all incidents analyzed, and provide a more complete view of anomalies for incidents that combine multiple types.

\vspace{-5pt}
\mypar{Root Cause in Event Data (Q2)}
Root causes are often interventions -- that is, explicit operations by actors that alter system resources in ways that trigger downstream issues. Those interventions are often observable in event data. These interventions vary in form: (1) In 32\% of incidents, the root cause can be attributed to a single actor performing an operation (e.g., resource deletion). In this case, a single root cause event is sufficient to explain the anomaly. (2) In 68\% of cases, root causes involve multiple actors or complex automated workflows (e.g., CI/CD pipelines), and intervention chains consist of many mutating events. Although the root cause can still be captured potentially by a single event, fully understanding it requires tracing through dependent services. (3) Common root causes include resource deletions, misconfigured deployments, and infrastructure updates.

The \zrhEvent{} incident presented in Table~\ref{tab:incidents-main} is complex, involving a series of events initiated by \rootcause{a code change}. The code change was merged into the main branch, which triggered an automated CI/CD pipeline, which in turn updated the infrastructure stack and \rootcause{improperly removed a critical \iam{} role \textit{(root cause)}}. As a result, many dependencies lost access to the removed role, leading to \anomaly{service failures in the sign-in website \textit{(anomaly)}}. It took the operators 2 hours to localize the \begin{wrapfigure}{r}{0.59\textwidth}
\vspace{-16pt}
\begin{tcolorbox}[left=2pt,right=2pt,top=0pt,bottom=0pt,
  enhanced,
  drop shadow={shadow xshift=1ex, shadow yshift=-1ex, opacity=0.3}]
{\small \textbf{Q2: How do root causes leave traces in event data?}}

\small \textit{\textbf{A:} 
Root causes frequently manifest as one or more events representing interventions on critical resources. These may be directly attributable to a single event (e.g., a deletion action) or distributed across a causal chain of actions (e.g., deployments).}
\end{tcolorbox}
\vspace{-20pt}
\end{wrapfigure}root cause. The incident was mitigated through a manual reactivation of the deleted role. This case highlights the need for interpretable RCL to provide information for detected anomalies, supporting faster incident mitigation.

\vspace{-5pt}
\subsection{Challenges}

Our empirical analysis reveals three major challenges in detecting anomalies and diagnosing their root causes from event data in cloud-based service systems:

\vspace{-5pt}
\mypar{C1. Overwhelming Event Volume}
The volume of events within a short time window can be overwhelming, making manual analysis difficult. Figure~\ref{fig:real-world-insights}b shows that in over 50\% of incidents, more than one thousand events needed to be examined, and this number could reach up to one billion. This scale stems from multiple services (Figure~\ref{fig:real-world-insights}c gives the number of services involved per incident) and multiple teams (Figure~\ref{fig:real-world-insights}d presents the number of teams involved per incident) generating concurrent events, significantly increasing the effort required to localize the root cause. For 71\% of incidents, identifying the root cause took more than 10 hours (Figure~\ref{fig:real-world-insights}e). This delay prevents timely mitigation. The financial impact can be severe as 81\% of incidents cost over \$1{,}000.

\vspace{-5pt}
\mypar{C2. Lack of Automatic Event-Based ADL}
The detection of incidents largely depends on alerts from impacted downstream services or user reports (93\% of incidents). The detection of incidents using event data (7\%) typically relies on time-consuming manual efforts from human operators. There is a clear need for proactive, event-based mechanisms that can detect and localize issues before they significantly affect downstream services or user experience.

\vspace{-5pt}
\mypar{C3. Lack of Interpretability and Actionability}
Even when alarms are triggered from downstream services, they often lack actionable insights. Most systems raise alarms without explaining what  went wrong (see Anomaly Symptoms in Table~\ref{tab:incidents-main}). Operators then need to search through large volumes of event data from different sources to find the root causes. There is a strong
\begin{wrapfigure}{r}{0.6\textwidth}
\vspace{-15pt}
\begin{tcolorbox}[left=2pt,right=2pt,top=0pt,bottom=0pt,
  enhanced,
  drop shadow={shadow xshift=1ex, shadow yshift=-1ex, opacity=0.3}]
\small
\textbf{Summary:} \textit{Real-world incident analysis highlights the need for an automated, scalable, and interpretable ADL framework. It motivates the design of \toolname{}, which detects anomalies by modeling event semantics and frequency, and localizes root causes by tracing recent interventions from detected anomalies.}
\end{tcolorbox}
\vspace{-15pt}
\end{wrapfigure}
need for an ADL framework that not only detects anomalies but also provides interpretable and actionable outputs such as root causes, so that operators can understand incidents and respond effectively.

\vspace{-5pt}
\section{\toolname{}: Our Proposed Framework} \label{sec:method}

We present \toolname{}, an open-box ADL framework for event data in cloud-based service systems. \toolname{} detects three anomaly types (\textit{Event Type}, \textit{Event Value}, and \textit{Event Frequency}), and provides root causes by tracing the causal structure of recent system interventions and detected anomalies.

\vspace{-5pt}
\subsection{Framework Overview}

\begin{figure*}
\centering
\includegraphics[width=\linewidth]{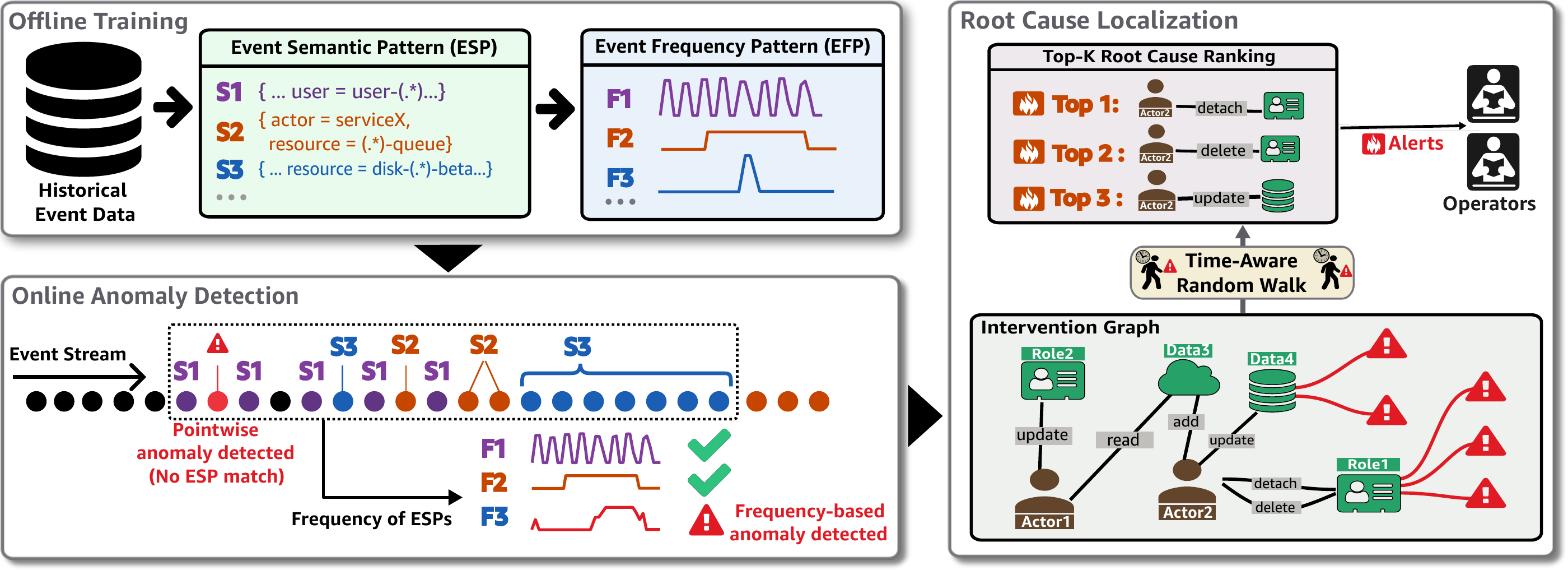}
\vspace{-20pt}
\caption{\textbf{Overview of \toolname{}}. Three phases: offline training (upper left), online anomaly detection (lower left), and RCL (right). Offline, \toolname{} learns ESPs and EFPs from historical events. Online, events are evaluated against these patterns: ESP identifies pointwise anomalies, while EFP detects frequency-based anomalies. Upon detection, \toolname{} constructs an Intervention Graph encoding causal links between interventions and anomalies, then applies a time-aware random walk to rank root causes.}
\label{fig:overview-method}
\vspace{-15pt}
\end{figure*}

\toolname{} operates in three phases: (1) Offline Training, (2) Online Anomaly Detection, and (3) RCL, as shown in Figure~\ref{fig:overview-method}. Each phase operates as follows:

\vspace{-5pt}
\mypar{Offline Training}
During offline training, \toolname{} learns two types of interpretable patterns from historical event data. (1) \textit{Event Semantic Patterns (ESPs)} capture semantic relationships across event fields (e.g., \texttt{actor}, \texttt{operation}, \texttt{resource}, and \texttt{time}) to represent known behaviors (e.g., normal interactions between system entities). (2) \textit{Event Frequency Patterns (EFPs)} model the frequency of these semantic patterns over time, enabling the detection of frequency-based anomalies. Together, ESPs and EFPs form the foundation for interpretable anomaly detection. 

\vspace{-5pt}
\mypar{Online Anomaly Detection}
As new events arrive, they are continuously compared against the learned ESPs and EFPs. An event is flagged as a \textit{pointwise anomaly} if it does not match any ESP (i.e., an Event Type or Event Value anomaly).  A time-window is flagged as a \textit{frequency-based anomaly} if the frequency of ESP-matching events deviates significantly from the corresponding EFP (i.e., an Event Frequency anomaly). These mechanisms capture the three anomaly types observed in practice. Although Event Order anomalies are defined in prior work~\cite{van2004workflow}, we found that they are exceedingly rare in practice and often manifest indirectly through frequency anomalies (Section~\ref{sec:analysis-real-world}).

\vspace{-5pt}
\mypar{Root Cause Localization}
When an anomaly is detected, \toolname{} constructs an \textit{Intervention Graph}, which encodes causal relationships between recent system interventions and the detected anomalies. \toolname{} then performs a time-aware random walk over the graph to identify the root cause(s), i.e., the intervention(s) most likely to be responsible for the detected anomalies. By RCL, \toolname{} provides interpretable and actionable insights, in contrast to prior approaches~\cite{event_ijcai16_ape, zengy2022shadewatcher, coskun2022detecting}, which only produce anomaly scores and require further manual investigation.

\begin{wrapfigure}{r}{0.59\textwidth}
\centering
\vspace{-30pt}
\begin{lstlisting}[language=json, basicstyle=\ttfamily\scriptsize, backgroundcolor=\color{gray!10}, frame=single, xrightmargin=5pt]
{"and": [
 {"==": [{"var": "actor.user.name"}, "merlinary"]},
 {"==": [{"var": "api.operation"}, "UpdateInstances"]},
 {"like": [{"var": "cloud.region"}, "^us-east-[1-4]$"]}
]}
\end{lstlisting}
\vspace{-7pt}
\caption{An ESP in the jsonLogic~\cite{jsonlogic} schema.}
\vspace{-10pt}
\label{fig:event-pattern}
\end{wrapfigure} 
\subsection{Event Semantic Pattern} \label{sec:esp}

Event Semantic Pattern (ESP) is a model of the normal event types and values observed in historical event data. They serve two purposes: a set of ESPs is used to detect \emph{pointwise anomalies} (i.e., single events that are anomalous on their own), and they allow labeling normal events with a specific ESP, which is then used to compute EFPs (Section~\ref{sec:efp}). \textit{Event Type} and \textit{Event Value} anomalies are pointwise anomalies; we must design ESPs to capture both of those types of anomalies. As for any other component of our system, we also require the ESP model to be interpretable and scalable.

Formally, the model learned by \toolname{} on a set of events is a set \(\mathcal{S} = \{s_1, s_2, \dots, s_k\}\) of ESPs, where each ESP \(s_i\) is an event-matching rule that captures expected behavior observed in the system. During online detection, if an incoming event \(e_j \in \mathcal{E}\) matches any pattern \(s_i \in \mathcal{S}\), it is labeled as normal with pattern \(s_i\); otherwise, it is flagged as an anomaly. A rule-based approach provides \emph{interpretability}, \emph{scalability} (Section~\ref{sec:eval-efficiency}), and \emph{deterministic results}.

ESPs are interpretable event matching expressions that capture the events observed in historical data. Figure~\ref{fig:event-pattern} shows an ESP that captures events where actor \texttt{\textcolor{NavyBlue}{merlinary}} performs operation \texttt{\textcolor{NavyBlue}{UpdateInstances}} across regions \texttt{\textcolor{NavyBlue}{us-east-[1-4]}}. During the online anomaly detection phase, an event with actor \texttt{\textcolor{NavyBlue}{merlinary}} performing \texttt{\textcolor{NavyBlue}{UpdateInstances}} in \texttt{\textcolor{NavyBlue}{us-}\textcolor{DarkRed}{west}\textcolor{NavyBlue}{-1}} is a pointwise anomaly, as no similar action has been observed before (assuming there are no other patterns matching it). Existing pattern synthesis techniques such as HyGLAD~\cite{hyglad} or log parsing techniques such as Drain~\cite{he2017drain} can extract patterns from event data. In the context of event-based ADL, we show that HyGLAD has a key advantage over Drain: it leverages the structured information in events directly to  take into account entity relationships. Table~B1 in the supplementary material gives an example where relationship-agnostic methods may mistakenly classify an abnormal event as normal, e.g., \texttt{\{user:\tinycolorbox{gray!20}{dev-1ac}, operation:\tinycolorbox{gray!20}{update}, resource:\tinycolorbox{gray!20}{prod-db-1}\}}. In our experiments, we use HyGLAD to learn ESPs in \toolname{}, and perform an ablation study using Drain (Section~\ref{sec:drain-study}).

\subsection{Event Frequency Pattern} \label{sec:efp}
Event Frequency Pattern (EFP) detects anomalies in \textit{Event Frequency}, which account for 67\% of the analyzed cases (Section~\ref{sec:analysis-real-world}). From our analysis, we observe that event frequencies are heterogeneous, as they may be dense (e.g., data transfers between services) or sparse (e.g., secret rotation events). We also require EFP to be interpretable and consistent with the open-box design of \toolname{}. 

Our EFP is inspired by the Matrix Profile~\cite{lu2022matrix}, which provides efficient and interpretable means to characterize frequency patterns through subsequence comparison. However, a key limitation of prior work is its emphasis on the \textit{shape} of subsequences, typically ignoring \textit{magnitude} differences~\cite{lu2022matrix}. While shape-based matching may be effective for continuously-valued signals, we argue that \textit{magnitude} is more important in the context of event-based anomaly detection. For instance, a known event pattern occurring at 1–3 requests per second (rps) may appear normal even at 5 rps, but a sudden spike to 100 rps is clearly anomalous. Shape-based anomaly detection methods~\cite{lu2022matrix} may fail to detect such deviations (Section~\ref{sec:ablation-shape-mag}). To address this, \toolname{} uses the Euclidean distance to directly compare subsequences, which is more suitable for event data. EFP is the first adaptation of the Matrix Profile for discrete time series derived from event data.

\noindent
\begin{minipage}[t]{0.58\textwidth}
\vspace{-5pt}
We construct a set of EFPs for a training period \([t_0, T]\) for each ESP \(s_i \in \mathcal{S}\). Let \(x^{(i)}_\tau\) denote the frequency of \(s_i\) at timestamp \(\tau\). Given a subsequence length \(M\), the set of all length-\(M\) sliding windows is defined as:
{\footnotesize \[
W^{(i)} = \left\{ w_u^{(i)} = \bigl(x_u^{(i)}, \dots, x_{u+M-1}^{(i)}\bigr) \in \mathbb{N}_0^M \;\middle|\; t_0 \le u \le T - M + 1 \right\}.
\]}

For each window \(w^{(i)}_u\), we compute the minimum Euclidean distance to a non-overlapping window \(w^{(i)}_v\), where \(\lvert u-v\rvert \ge M\):
\begin{equation}  
\footnotesize
d^{(i)}_u \;=\; \min_{\substack{v = t_0,\dots,T-M+1 \\ |u-v| \ge M}}  
\;\bigl\lVert w^{(i)}_u - w^{(i)}_v \bigr\rVert_2.
\end{equation}

We use \(f_i\) to denote the EFP with respect to the ESP \(s_i \in \mathcal{S}\) as the sequence of nearest distances computed from \(W^{(i)}\),
\begin{equation} \label{eq:fi}
\footnotesize
f_i = \left\{ d_u^{(i)} \right\}_{u=t_0}^{T - M + 1}.
\end{equation}

\end{minipage}
\hfill
\begin{minipage}[t]{0.40\textwidth}
\centering
\vspace{-7pt}
\includegraphics[width=\linewidth]{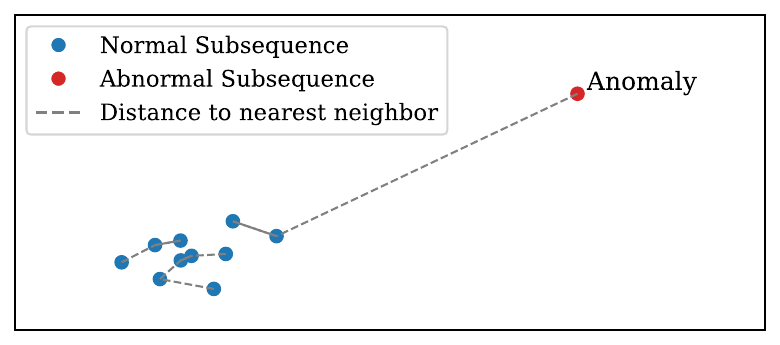}
\vspace{-20pt}
\captionof{figure}{Detecting anomalies with EFP. Each subsequence \(w_u^{(i)}\) is a point in Euclidean space and linked to its nearest non-trivial neighbor \(w_v^{(i)}\). The set of distances \(\{d_u^{(i)}\}\) forms the Event Frequency Pattern (EFP) \(f_i \in \mathcal{F}\). The \textcolor{DarkRed}{abnormal subsequence} lies far from the cluster of \textcolor{NavyBlue}{normal subsequences}, as it has a \textit{statistically} large distance to its nearest neighbor, indicating a potential anomaly.}
\label{fig:efp}
\vspace{3pt}
\end{minipage}

The set of all EFPs is denoted by \(\mathcal{F} = \{ f_1, f_2, \dots, f_{|\mathcal{S}|} \}\). Each EFP \(f_i \in \mathcal{F}\) provides an \textit{interpretable} measure of self-similarity, showing how typical each subsequence is when compared to others (see Figure~\ref{fig:efp}). An anomaly naturally arises when a subsequence has an abnormally large distance to its nearest neighbor, indicating that the observed frequency is \textit{very} different from all known patterns.

To conclude if \(w^{(i)}_{\mathrm{new}}\) is anomalous, we perform hypothesis testing on whether \(d^{(i)}_{\mathrm{new}}\) is statistically consistent with \(f_i\). Specifically, we treat \( f_i \) as samples drawn from an unknown distribution \( P_D \), representing normal frequency of \( s_i \). Then, we define the empirical cumulative distribution function over \(f_i\) as \(\hat{F}_D(x) = \frac{1}{|f_i|} \sum_{d \in f_i} \mathbf{1}\{d \le x\}\). The empirical survival function is then given by \( \hat{S}(d_{\mathrm{new}}) = 1 - \hat{F}_D(d_{\mathrm{new}}), \) which quantifies the probability of observing a distance as large as \( d_{\mathrm{new}} \). Let \( \alpha \) be a chosen significance level (e.g., \( \alpha = 0.01 \)). If \( \hat{S}(d_{\mathrm{new}}) < \alpha \), we reject the null hypothesis that \( d_{\mathrm{new}} \sim P_D \) and classify the window as anomalous.  This design is distribution-free, making it well-suited to the heterogeneous event frequencies found in large-scale cloud systems.

\noindent \textbf{Robustness to Noise.}
A key challenge in real-world anomaly detection is the presence of noise (e.g., unknown anomalies) in training data. Many prior methods~\cite{hyglad, aydore2022detecting, event_ijcai16_ape, du2017deeplog} assume that training data is anomaly-free, which is often unrealistic. In contrast, our approach is robust to such noise because hypothesis testing treats these cases as statistically insignificant. As long as they do not dominate the empirical distribution during normal periods, they have little impact on overall performance. We empirically validate this in Section~\ref{sec:robustness}.

\noindent \textbf{Adaptation.} 
Cloud systems are constantly evolving with new behaviors, and without adaptation, new behaviors would be repeatedly flagged as anomalies. \toolname{}'s adaptive mechanism works as follows: when an event does not match any ESP, \toolname{} first flags it as an anomaly then records its count \(c_s\) per window. If \(c_s\) exceeds a threshold \(T_s\) within \(N\) successive windows (i.e., persistent), we invoke HyGLAD to revise the ESP set, ensuring only persistent behaviors are added as new normals. For EFPs, \toolname{} continuously updates \(f_i\) with newly observed normal distances \(d_u^{(i)}\) (Equation~\ref{eq:fi}), keeping hypothesis testing (Section~\ref{sec:efp}) aligned with evolving system characteristics. We note that no online detector can instantly distinguish legitimate system changes from true anomalies without external context (e.g., release notes). Therefore, we also localize root causes to provide interpretable explanations of why the detected anomalies occur (Section~\ref{sec:anomaly-localization}).

\begin{wrapfigure}{r}{0.44\linewidth}
\vspace{-40pt}
\centering
\scriptsize
\begin{minipage}{\linewidth}
\begin{algorithm}[H]
\captionsetup{font=scriptsize}
\caption{Construction of the Intervention Graph}
\label{alg:construct-intervention-graph}
\begin{algorithmic}[1]
\Require Event set $\mathcal{E} = \{e_1, e_2, \dots, e_n\}$
\Ensure Intervention graph $\mathcal{G} = (\mathcal{V}, \mathcal{E}_G)$
\State Initialize graph $\mathcal{G} \gets (\mathcal{V} \gets \emptyset, \mathcal{E}_G \gets \emptyset)$
\For{each event $e_i \in \mathcal{E}$}
    \State Extract \texttt{actor}, \texttt{operation}, \texttt{resources}, \texttt{time} from $e_i$
    \State Add \texttt{actor} node to $\mathcal{V}$ if not already present
    \For{each \texttt{resource} $r_j$ in \texttt{resources}}
        \State Add node $r_j$ to $\mathcal{V}$ if not already present
        \State Add edge $(\texttt{actor} \xrightarrow{\texttt{operation}, \texttt{time}} r_j)$ to $\mathcal{E}_G$
        \If{$e_i$ is associated with an anomaly}
            \State Add \texttt{Anomaly} node to $\mathcal{V}$ if not already present
            \State Add edge $(r_j \xrightarrow{\texttt{time}} \texttt{Anomaly})$ to $\mathcal{E}_G$
        \EndIf
    \EndFor
\EndFor
\State \Return $\mathcal{G}$
\end{algorithmic}
\end{algorithm}
\end{minipage}
\vspace{-10pt}
\end{wrapfigure}
\subsection{Root Cause Localization} \label{sec:anomaly-localization}
\toolname{} localizes the root cause to provide interpretable explanations of what triggered the anomaly and why it occurred. Our empirical analysis reveals that many incidents stem from improper interventions that modify one or more system resources, thus leading to anomalies observable in different parts of the system (see Section~\ref{sec:analysis-real-world}). Motivated by this observation, we introduce the concept of an \textit{Intervention Graph}, which encodes the temporal and causal relationships between system interventions and detected anomalies. This graph can be constructed directly from event data and enables automatic RCL.

\vspace{-5pt}
\subsubsection{Intervention Graph}
Given a set of structured events $\mathcal{E} = \{e_1, e_2, \ldots, e_n\}$ within a time window $\mathcal{W}_{t}'$, the \emph{Intervention Graph} is a directed multigraph $\mathcal{G} = (\mathcal{V}, \mathcal{E}_G)$ where $\mathcal{V}$ is the set of nodes representing actors, resources, and anomalies; and $\mathcal{E}_G$ is the set of directed labeled edges representing event-level interactions. For each event $e_i$ involving actor $\Lambda_i$, operation $\Omega_i$, time $\tau_i$, and resource set $\{r_1, \ldots, r_k\}$, we add directed edges from $\Lambda_i$ to each $r_j$, labeled with $(\Omega_i, \tau_i)$. If $e_i$ is anomalous, we add an edge from each $r_j$ to an anomaly node, labeled with $\tau_i$ (see Algorithm~\ref{alg:construct-intervention-graph}).

The resulting graph provides a unified view of how interventions propagate and impact system components (see Figure~\ref{fig:overview-method}). It enables operators to visually trace anomaly paths and interpret potential root causes. Nevertheless, in practice, large-scale systems generate complex graphs that are difficult to inspect manually. To this end, we introduce a time-aware random walk algorithm that ranks root cause candidates based on traversal frequency and temporal plausibility.

\begin{wrapfigure}{R}{0.43\linewidth}
\vspace{-25pt}
\centering
\scriptsize
\begin{minipage}{\linewidth}
\begin{algorithm}[H]
\captionsetup{font=scriptsize}
\caption{Time-Aware Random Walk} \label{alg:temporal-random-walk}
\begin{algorithmic}[1]
\Require Graph $G$, anomaly nodes $A$, no. walks $N$
\Ensure Node-level \(visitCount\)

\State Initialize $visitCount \gets \{\}$ 
\ForAll{$(a, t_a) \in A$}
    \For{$i = 1$ to $N$}
        \State $u \gets a$, $t \gets t_a$
        \While{\textbf{true}}
            \State $preds \gets$ Predecessors of $u$ in $G$
            \State $validEdges \gets []$
            \ForAll{$p \in preds$}
                \ForAll{edges $(p, u, e)$ in $G$}
                    \State $t_e \gets e.\text{event\_time}$
                    \If{$t_e \le t$}     \Comment{\textcolor{NavyBlue}{Time-Aware}}
                        \State Append $(p, t_e)$ to $validEdges$
                    \EndIf
                \EndFor
            \EndFor
            \If{$validEdges = \emptyset$}
                \State \textbf{break}
            \EndIf
            \State Sample $(p', t')$ from $validEdges$
            \State $visitCount[p'] \gets visitCount[p'] + 1$
            \State $u \gets p'$, $t \gets t'$
        \EndWhile
    \EndFor
\EndFor
\State \Return $visitCount$
\end{algorithmic}
\end{algorithm}
\end{minipage}
\vspace{-10pt}
\end{wrapfigure}
\vspace{-5pt}
\subsubsection{Time-Aware Random Walk} \label{sec:random-walk}

To automate RCL, we apply a random walk algorithm over the Intervention Graph. The walk begins at each anomaly node and traverses backward through the graph toward potential sources of the anomaly. Nodes visited frequently across multiple walks are identified as root causes. The intuition behind this design is that true root causes often have multiple causal paths leading to observed anomalies, so the random walker will visit them more frequently than unrelated nodes. This assumption is empirically grounded and used in~\cite{salesforce_causalai23,rcd,pham2024root,xin2023causalrca}. Our analysis (Section~\ref{sec:analysis-real-world}) shows that root causes are often interventions that alter system resources, triggering downstream anomalies through multiple causal paths. Our Intervention Graph captures these paths directly from event data, allowing the random walk to visit the root cause frequently from the detected anomalies.

However, a naive traversal may violate causal time constraints. For instance, tracing from an anomaly at 11:00 AM to an operation that occurred at 2:00 PM creates an invalid path. We propose to use a \textit{time-aware random walk} that restricts movement to edges where the associated event occurred no later than the current timestamp within \(\mathcal{W}_{t}'\). This ensures that all paths are temporally valid (see Algorithm~\ref{alg:temporal-random-walk}).

\vspace{-10pt}
\subsubsection{Root Cause Ranking}
\toolname{} returns a ranked list of root causes (i.e., $\texttt{actor} \xrightarrow{\texttt{operation}, \texttt{time}} \texttt{resource}$), where the ranking is determined by the visit counts from the time-aware random walk. Interventions that are more frequently visited along valid temporal paths from anomalies are considered more likely to be the root causes. Given this ranked list, operators can focus on the top-ranked interventions instead of manually inspecting all events. Furthermore, the sub-Intervention Graph associated with each ranked root cause can be visualized (Figure~\ref{fig:overview-method}), reinforcing our open-box design. These visualizations provide interpretable, path-based explanations of how anomalies may have propagated from specific actors or resources through chains of operations.

\vspace{-5pt}
\section{Experiments} \label{sec:experiments}

We conduct extensive experiments to answer the following questions:

\begin{itemize}[leftmargin=*, topsep=0pt, itemsep=0pt]
\item RQ1: How effective is \toolname{} in detecting anomalies?
\item RQ2: How effective is \toolname{} in localizing root causes?
\item RQ3: How efficient is \toolname{}?
\item RQ4: What is the contribution of each component to \toolname{}?
\item RQ5: How robust is \toolname{}?
\end{itemize}

\vspace{-8pt}
\subsection{Benchmark Datasets}\label{sec:dataset}

At the time of conducting this research, there is no publicly available benchmark for event-based ADL in cloud systems with annotated anomalies and their corresponding root causes. To address this gap, we construct \textbf{five benchmark datasets}: three by reproducing incidents on real-world cloud service systems, and two collected from \textbf{historical incidents}. We describe both the incident reproductions on benchmark systems and the collected historical incidents in detail below.

\vspace{-5pt}
\subsubsection{Benchmark Systems} 
We use three service systems (Falcon, Flask and Live) deployed on real infrastructure to reproduce incidents and collect event data. \textbf{Falcon} is a cloud-based web application platform consisting of 363 actors and 138{,}292 resources, designed to host websites. \textbf{Flask} is a microservice-based music catalog and retrieval system, comprising 531 actors and 143{,}353 resources. \textbf{Live} is a real-time cricket scoring platform serving 2{,}507 actors and managing 1{,}404 resources. Each system generates events across multiple layers (e.g., API calls, configurations, and resource updates). We reproduced three common types of incidents identified in our empirical study. (1)~\textit{Secret Deactivation}: We randomly deactivate a secret (e.g., access key), causing permission errors across dependent services and triggering anomalies in event types and frequencies. The root cause is the deactivation event. (2)~\textit{Denial-of-Service (DoS)}: We flood API calls, leading to system-wide throttling. Anomalies manifest as spikes in request rates and failure events. The root cause is the actor repeatedly performing the DoS operations. (3)~\textit{Unusual Activity}: We reproduce a compromised credential scenario by creating a random resource in an unusual region. These unusual operations are simultaneously anomalies and root causes. We randomly repeat these actions across different resources and permission settings, yielding 30 one-hour test samples per system.

\vspace{-5pt}
\subsubsection{Real-world Incidents} 
We collected data from two incidents in 2024 at \aws{}: \textbf{\zrhEvent{}} and \textbf{\mxpEvent{}} (Table~\ref{tab:incidents-main}). The incident \textbf{\zrhEvent{}} was reported after customers were unable to log in to the \aws{} Management Console or switch between service consoles in Region A for 1h49m. Around 33.1\% of the requests failed due to 4xx/5xx errors. The issue stemmed from an improper deletion of a critical role during a code deployment via an infrastructure pipeline. Diagnosing the incident took 100 minutes and involved multiple teams and services. The dataset includes 26{,}018 actors and 249 resources. \toolname{} learned 273 ESPs from the normal operation period, reflecting the complex operational patterns in the production environment. The incident \textbf{\mxpEvent{}} was reported after 3{,}355 \aws{} accounts in Region B experienced ServiceD API failures for 34 minutes, affecting 89 services in total. The root cause was an incorrect deactivation of an access key, due to a software defect and its recent deployments. The issue was resolved by reactivating the key. \toolname{} learned 308 ESPs for this dataset, capturing the diverse interaction patterns across the affected services.

\vspace{-5pt}
\subsection{Evaluation Metrics}

\vspace{-2pt}
\subsubsection{Anomaly Detection}
Following existing work~\cite{pham2024baro, hyglad},
we use Precision, Recall, and F1 scores to evaluate the anomaly detectors. When an anomaly detection algorithm successfully detects an abnormal sample (i.e., a case with anomalies), the detection is counted as a True Positive (TP). Conversely, incorrectly classifying an abnormal sample as normal is considered False Negative (FN). Likewise, incorrectly classifying a normal sample as abnormal is considered False Positive (FP).
{\bf Precision} is the ratio \(\frac{TP}{TP + FP}\), {\bf Recall} is \(\frac{TP}{TP + FN}\) and the {\bf F1-score} is determined by \(\frac{2\times Precision \times Recall}{Precision + Recall}\).

\vspace{-2pt}
\subsubsection{Root Cause Localization}
We use two standard metrics: $AC@k$ and $Avg@k$ to assess the root cause localization (RCL) performance~\cite{pham2024baro, circa}. Herein, we set $k = 1, 3, 5$. Given a set of failure cases A, \(AC@k\) is determined by \(\frac{1}{|A|} \sum\nolimits_{a\in A}\frac{\sum_{i<k}R^a[i]\in V^a_{rc}}{min(k, |V^a_{rc}|)}\), and then \(Avg@k\) is calculated by \(\frac{1}{k}\sum_{j=1}^k AC@j\), where $R^a[i]$ denotes the $i$th ranking result for the failure case $a$ by an RCL method, and $V^a_{rc}$ is the true root cause set of case $a$. $AC@k$ represents the probability the top $k$ results given by a method include the real root causes. $Avg@k$ measures the overall performance of RCL methods.

\vspace{-5pt}
\subsection{Baselines}

We evaluate \toolname{} against two categories: (i) anomaly detection, and (ii) RCL.  
For anomaly detection, we include:
ADAMAS~\cite{gu2025adamas},
APE~\cite{event_ijcai16_ape},
BARO~\cite{pham2024baro},
CUSUM~\cite{cusum},
DeepSVDD~\cite{deepsvdd},
DIF~\cite{dif},
ICL~\cite{icl},
KPIRoot~\cite{gu2024kpiroot},
NeuralLog~\cite{le2021neurallog},
NeuTraL~\cite{neutral},
NSigma~\cite{circa},
RCA~\cite{rca},
RDP~\cite{rdp}, and
ShadeWatcher~\cite{zengy2022shadewatcher}.  
For RCL, we compare against:
BARO~\cite{pham2024baro},
CausalAI~\cite{salesforce_causalai23},
CausalRCA~\cite{xin2023causalrca},
DeepHunt~\cite{sun2025interpretable},
\(\epsilon\)-Diagnosis~\cite{ediagnosis},
Groot~\cite{Wang2021Groot},
KPIRoot~\cite{gu2024kpiroot},
RCD~\cite{rcd},
TVDiag~\cite{xie2025tvdiag},
and a \textit{Random} baseline that selects a root cause at random. We use available implementations when possible. For Groot, APE, and ShadeWatcher, we implement them  based on the algorithmic details provided in the original papers. For ADAMAS, DeepHunt, TVDiag, and KPIRoot, we adapt the publicly available code to work with event data. ADAMAS, originally designed for monitoring metrics, is adapted to operate on frequency-based time series derived from ESPs. KPIRoot uses both similarity and causality analysis on the same time series representation. DeepHunt and TVDiag, designed for multimodal data (logs, traces, metrics), are adapted by constructing intervention graphs and feature representations from event data. All hyperparameters follow the recommendations in the respective papers. For methods with configurable thresholds, we follow established practice by running multiple settings and reporting the best performance. Due to space constraints, the detailed description of these baselines are in supplementary material in our replication package.

\subsection{Experimental Setting}

We conduct all the experiments on Linux servers equipped with 12 CPU and 36\,GB RAM. To manage randomness, we repeat each experiment ten times, then report the mean and standard deviation of the collected results. We give each method two hours to finish their tasks, otherwise a \textit{time-out} error is thrown. Our framework is implemented using Python 3.12.

\begin{table}[t]
\caption{The anomaly detection performance of \toolname{} and fourteen baselines on five datasets (Falcon, Flask, Live, \zrhEvent{}, \mxpEvent{}) in terms of Precision, Recall, and F1-score. We report the mean and standard deviation over ten runs with different random seeds. We \textbf{bold} the best values and \underline{underline} the second-best.}
\label{tab:rq1-anomaly-detection}
\vspace{-10pt}
\resizebox{\textwidth}{!}{%
\setlength\tabcolsep{1pt}
\begin{tabular}{lccccccccccccccc}
\toprule
\multirow{2}{*}{\textbf{Method}} 
 & \multicolumn{3}{c}{\textbf{Falcon}} 
 & \multicolumn{3}{c}{\textbf{Flask}} 
 & \multicolumn{3}{c}{\textbf{Live}} 
 & \multicolumn{3}{c}{\textbf{\zrhEvent{}}} 
 & \multicolumn{3}{c}{\textbf{\mxpEvent{}}} \\ 
 \cmidrule(lr){2-4} \cmidrule(lr){5-7} \cmidrule(lr){8-10} \cmidrule(lr){11-13} \cmidrule(lr){14-16}
& Precision & Recall & F1-Score & Precision & Recall & F1-Score & Precision & Recall & F1-Score & Precision & Recall & F1-Score & Precision & Recall & F1-Score \\ \midrule
\rowcolor{linegray} APE~\cite{event_ijcai16_ape} & 0.46±0.15 & 0.35±0.16 & 0.38±0.12 & 0.47±0.18 & 0.68±0.28 & 0.50±0.09 & 0.36±0.21 & 0.55±0.28 & 0.37±0.15 & 0.16±0.07 & \textbf{1.00±0.00} & 0.27±0.10  & \underline{0.25±0.09} & 0.98±0.04 & \underline{0.39±0.11} \\ 
BARO~\cite{pham2024baro} & 0.18±0.00 & \textbf{1.00±0.00} & 0.30±0.00 & 0.17±0.00 & \textbf{1.00±0.00} & 0.29±0.00 & 0.17±0.00 & \textbf{1.00±0.00} & 0.29±0.00 & 0.06±0.00 & \textbf{1.00±0.00} & 0.12±0.00 & 0.12±0.00 & \textbf{1.00±0.00} & 0.22±0.00 \\ 
\rowcolor{linegray} CUSUM~\cite{cusum} & 0.53±0.00 & \underline{0.84±0.00} & 0.65±0.00 & 0.71±0.00 & \textbf{1.00±0.00} & \underline{0.83±0.00} & \underline{0.77±0.00} & \textbf{1.00±0.00} & \underline{0.87±0.00} & 0.06±0.00 & \textbf{1.00±0.00} & 0.12±0.00 & 0.12±0.00 & \textbf{1.00±0.00} & 0.22±0.00 \\ 
DeepSVDD~\cite{deepsvdd} & 0.62±0.45 & 0.39±0.38 & 0.44±0.40 & 0.47±0.46 & 0.34±0.34 & 0.34±0.36 & 0.70±0.44 & 0.37±0.34 & 0.43±0.36 & 0.82±0.10 & 0.81±0.24 & 0.79±0.12 & 0.15±0.04 & 0.80±0.21 & 0.25±0.05 \\
\rowcolor{linegray} DIF~\cite{dif} & 0.15±0.17 & 0.27±0.41 & 0.10±0.15 & 0.21±0.26 & 0.37±0.38 & 0.19±0.20 & 0.18±0.28 & 0.24±0.37 & 0.15±0.23 & 0.43±0.12 & 0.84±0.24 & 0.53±0.03 & 0.10±0.01 & \underline{0.99±0.01} & 0.18±0.01 \\
ICL~\cite{icl} & \underline{0.66±0.44} & 0.39±0.38 & 0.43±0.40 & 0.54±0.49 & 0.36±0.38 & 0.32±0.35 & 0.61±0.49 & 0.22±0.27 & 0.29±0.32 & \underline{0.87±0.05} & \underline{0.99±0.01} & \textbf{0.92±0.03} & - & - & - \\
\rowcolor{linegray} NeuralLog~\cite{le2021neurallog} & 0.60±0.03 & \textbf{1.00±0.00} & \underline{0.75±0.02} & 0.42±0.18 & 0.39±0.21 & 0.33±0.09 & 0.53±0.01 & \textbf{1.00±0.00} & 0.69±0.01 & 0.06±0.00 & \textbf{1.00±0.00} & 0.12±0.00 &  0.15±0.00 & \textbf{1.00±0.00} & 0.26±0.00  \\ 
NeuTraL~\cite{neutral} & \textbf{0.82±0.34} & 0.55±0.39 & 0.60±0.38 & 0.68±0.45 & 0.45±0.38 & 0.49±0.39 & 0.68±0.46 & 0.35±0.36 & 0.41±0.37 & \textbf{0.89±0.03} & 0.91±0.21 & 0.89±0.14 & - & - & -\\
\rowcolor{linegray} NSigma~\cite{circa} & 0.37±0.00 & \textbf{1.00±0.00} & 0.54±0.00 & 0.17±0.00 & \textbf{1.00±0.00} & 0.29±0.00 & 0.17±0.00 & \textbf{1.00±0.00} & 0.29±0.00 & 0.07±0.00 & \textbf{1.00±0.00} & 0.13±0.00 & 0.16±0.00 & \textbf{1.00±0.00} & 0.28±0.00 \\ 
RCA~\cite{rca} & 0.40±0.39 & 0.37±0.41 & 0.26±0.30 & 0.30±0.37 & 0.46±0.41 & 0.28±0.31 & 0.27±0.37 & 0.32±0.40 & 0.22±0.30 & 0.55±0.21 & 0.62±0.22 & 0.53±0.08 & 0.14±0.02 & 0.45±0.12 & 0.21±0.04 \\
\rowcolor{linegray} RDP~\cite{rdp} & 0.14±0.16 & 0.80±0.40 & 0.21±0.23 & 0.12±0.26 & 0.80±0.40 & 0.15±0.28 & 0.12±0.27 & 0.50±0.50 & 0.14±0.28 & 0.11±0.00 & \textbf{1.00±0.00} & 0.20±0.00 & - & - & -\\
ShadeWat~\cite{zengy2022shadewatcher} & 0.17±0.02& 0.76±0.25 & 0.27±0.03 & 0.17±0.01 & \underline{0.82±0.20} & 0.28±0.03 & 0.17±0.02 & \underline{0.90±0.12} & 0.30±0.03 & 0.06±0.01 & 0.90±0.14 & 0.11±0.01 & 0.11±0.03 & 0.83±0.35 & 0.19±0.06 \\ 
\rowcolor{linegray} ADAMAS~\cite{gu2025adamas} & 0.19±0.00 & \textbf{1.00±0.00} & 0.31±0.00 & 0.18±0.00 & \textbf{1.00±0.00} & 0.30±0.00 & 0.17±0.00 & \textbf{1.00±0.00} & 0.29±0.00 & 0.06±0.00 & \textbf{1.00±0.00} & 0.12±0.00 & 0.12±0.00 & \textbf{1.00±0.00} & 0.22±0.00 \\
KPIRoot~\cite{gu2024kpiroot} & 0.61±0.00 & 0.71±0.00 & 0.66±0.00 & \underline{0.80±0.00} & 0.53±0.00 & 0.64±0.00 & 0.70±0.00 & 0.53±0.00 & 0.60±0.00 & 0.21±0.00 & 0.40±0.00 & 0.28±0.00 & 0.00±0.00 & 0.00±0.00 & 0.00±0.00 \\
\midrule
\rowcolor{lineblue} \textbf{\toolname{}} & \textbf{0.82±0.00} & \textbf{1.00±0.00} & \textbf{0.90±0.00} & \textbf{0.81±0.00} & \textbf{1.00±0.00} & \textbf{0.90±0.00} & \textbf{0.91±0.00} & \textbf{1.00±0.00} & \textbf{0.95±0.00} & 0.81±0.00 & \textbf{1.00±0.00} & \underline{0.90±0.00} & \textbf{0.91±0.00} & \textbf{1.00±0.00} & \textbf{0.95±0.00} \\ \midrule
\rowcolor{linegreen} \textit{ESP-only} & 0.72±0.00 & 1.00±0.00 & 0.84±0.00 & 0.37±0.00 & 1.00±0.00 & 0.54±0.00 & 0.31±0.00 & 1.00±0.00 & 0.47±0.00 & 0.72±0.00 & 1.00±0.00 & 0.84±0.00 & 0.91±0.00 & 1.00±0.00 & 0.95±0.00\\
\rowcolor{linegreen2} \textit{EFP-only} & 0.79±0.00 & 1.00±0.00 & 0.88±0.00 & 0.52±0.00 & 0.83±0.00 & 0.64±0.00 & 0.18±0.00 & 0.50±0.00 & 0.26±0.00 & 0.59±0.00 & 1.00±0.00 & 0.74±0.00 & 1.00±0.00 & 1.00±0.00 & 1.00±0.00 \\ 
\midrule
\rowcolor{linegreen} \textit{EventADL (C)} & 0.83±0.01 & 0.96±0.05 & 0.89±0.03 & 0.82±0.00 & 0.93±0.00 & 0.87±0.00 & 0.92±0.02 & 0.90±0.03 & 0.91±0.01 & 0.62±0.08 & 0.93±0.12 & 0.75±0.09 & 0.94±0.06 & 0.93±0.06 & 0.85±0.18 \\ 
\rowcolor{linegreen2} \textit{ESP-only (C)} & 0.73±0.03 & 0.78±0.05 & 0.76±0.04 & 0.37±0.06 & 0.81±0.13 & 0.51±0.08 & 0.34±0.03 & 0.88±0.04 & 0.49±0.03 & 0.93±0.06 & 0.70±0.17 & 0.78±0.10 & 0.97±0.06 & 0.87±0.06 & 0.83±0.16 \\
\rowcolor{linegreen} \textit{EFP-only (C)} & 0.65±0.05 & 0.83±0.05 & 0.72±0.05 & 0.59±0.07 & 0.78±0.09 & 0.67±0.08 & 0.17±0.01 & 0.40±0.03 & 0.24±0.01 & 0.67±0.05 & 0.87±0.06 & 0.75±0.01 & 1.00±0.00 & 0.83±0.21 & 0.86±0.20 \\ \midrule
\rowcolor{linegreen2}\textit{\toolname{} (D1)} & 0.53±0.00 & 1.00±0.00 & 0.70±0.00 & 0.51±0.00 & 1.00±0.00 & 0.67±0.00 & 0.88±0.00 & 0.93±0.00 & 0.90±0.00 & 0.62±0.00 & 1.00±0.00 & 0.77±0.00 & 0.91±0.00 & 1.00±0.00 & 0.95±0.00 \\ 
\rowcolor{linegreen} \textit{\toolname{} (D2)} & 0.72±0.00 & 0.94±0.00 & 0.82±0.00 & 0.75±0.00 & 1.00±0.00 & 0.86±0.00 & 0.97±0.00 & 0.93±0.00 & 0.95±0.00 & 0.43±0.00 & 1.00±0.00 & 0.61±0.00 & 0.83±0.00 & 1.00±0.00 & 0.91±0.00 \\ 
\bottomrule
\end{tabular}
}

{\tiny \textit{Note}: Methods shown in \tinycolorbox{linegreen}{\textit{italic}} are used in the ablation study discussed in Section~\ref{sec:ablation}. (-) ICL, NeuTraL, and RDP encounter OOM/time-out errors in the \mxpEvent{} dataset.}
\vspace{-10pt}
\end{table}

\subsection{RQ1: How effective is \toolname{} in detecting anomalies?}

In this RQ, we evaluate the performance of \toolname{} and the anomaly detection baselines across five datasets. We use the event data during the incident as anomalous samples and event data during normal operation as normal data. We report the average of Precision, Recall, and F1-score over all the cases. Table~\ref{tab:rq1-anomaly-detection} presents the experimental results, with the best results highlighted in \textbf{bold} and second-best results are \underline{underlined}. We draw the following observations:

\textbf{(1) \toolname{} is highly effective across all datasets, outperforming most baselines by large margins.} \toolname{} achieves the highest F1-score on 4 out of 5 datasets and consistently maintains strong performance with F1-score at least 0.90 on all datasets. Because ESPs and EFPs are designed to capture any deviation from training data, \toolname{} yields 100\% recall across benchmarks. The consistent precision above 0.8 shows that ESP and EFP also generalize well over their training data, avoid false positives. Finally, \toolname{} shows deterministic performance across runs (i.e., standard deviation = 0), since ESP and EFP training is not affected by randomness.

\textbf{(2) Metric-based methods achieve high recall scores but suffer from low precision.} Metric-based anomaly detectors such as BARO~\cite{pham2024baro}, CUSUM~\cite{cusum}, NSigma~\cite{circa}, ADAMAS~\cite{gu2025adamas}, and KPIRoot~\cite{gu2024kpiroot} do not generalize well with event-frequency time series, which exhibit different statistical properties than metrics time series. BARO employs Bayesian Online Change Point Detection for 
multivariate time series, while CUSUM and NSigma detect deviations in univariate time series based on the mean and standard deviation. These methods detect many anomalies because anomalies in cloud systems often affect multiple entities and manifest as frequency-based deviations (e.g., spikes in error events). BARO models only frequency dependencies with Bayesian statistics, whereas CUSUM and NSigma rely solely on Gaussian assumptions. Such simplifications hinder their ability to model dynamic and heterogeneous behaviors of cloud systems, making them less suitable for complex event-based time series. ADAMAS~\cite{gu2025adamas}, an adaptive AutoML framework that uses Bayesian Optimization to automatically select and tune its models, achieves 100\% recall but low precision. When adapted to event frequencies, its underlying models do not generalize well. KPIRoot~\cite{gu2024kpiroot} achieves moderate F1-scores on benchmark datasets and fails completely on the \mxpEvent{} dataset. KPIRoot uses R-space analysis and ratio-based scoring to detect anomalies in continuous time series like CPU usage and memory utilization. When applied to event-frequency time series with discrete event counts, KPIRoot does not generalize well, leading to inconsistent performance on complex real-world incidents. In contrast, our EFP uses a magnitude-based subsequence distance approach tailored to event-frequency patterns, achieving both high recall and precision.

\textbf{(3) Deep learning-based methods exhibit moderate performance and poor robustness.} Deep learning baselines such as DeepSVDD~\cite{deepsvdd}, DIF~\cite{dif}, ICL~\cite{icl}, and RCA~\cite{rca} achieve lower F1-score than \toolname{}. They fail to generalize normal event behavior. For instance, RCA trains an ensemble of autoencoders to reconstruct normal data patterns, but autoencoders may degenerate to identity mappings, thereby missing anomalies with low reconstruction loss. More broadly, these methods assume clean training data, which is rarely realistic--unknown anomalies in the training set can significantly  degrade performance. For example, DeepSVDD~\cite{deepsvdd} is a one-class classification method that learns to enclose normal data within a hypersphere in latent space, which may also incorrectly  generalize anomalies into the learnt hypersphere, leading to low recall as they cannot catch anomalies. In addition, deep learning methods are inherently stochastic, relying on random initialization and sampling, which yields slightly different performance across different runs. This can undermine engineers' trust, as repeated runs lead to inconsistent results. By contrast, \toolname{}'s open-box design and deterministic behavior ensure both accuracy and reliability.

\subsection{RQ2: How effective is \toolname{} in finding the root cause of anomalies?}

In this section, we evaluate the performance of \toolname{} against the existing baselines on all five datasets. We use all events within 1 hour around the anomaly occurrence time for all methods to perform RCL, as we have observed in our benchmark cloud systems and in the collected real-world incidents that the root causes are closely aligned with the anomaly occurrence time. As the cloud systems are highly dynamic, a mistaken operation propagates and affects the systems quickly. We also use the Random baseline as a proxy for a human operator that randomly generates a hypothesis and investigates the root cause, aiming to see how RCL can help increase root cause analysis performance. Table~\ref{tab:rq2-rca} reports the RCL performance of all methods using the AC@1, AC@3, and Avg@5 scores across all five datasets. We run each experiment 10 times with different random seeds and report the mean and standard deviation of their performance. We observe that:

\textbf{(1) \toolname{} consistently outperforms all baselines in RCL}. It achieves an AC@3 score of 100\% on all systems, meaning the true root cause is always ranked among the top three candidates. This result validates our assumption that root causes have multiple causal paths to observed anomalies (Section~\ref{sec:random-walk}), ensuring the random walker reaches the true root cause. In the Live, \zrhEvent{}, and \mxpEvent{} datasets, \toolname{} can rank the root causes precisely in the top-1 ranking, achieving an AC@1 of 100\%. In the Falcon and Flask datasets, there are a few cases where \toolname{} could not rank the true root cause (i.e., the actor and their interventions) as the top-1 root cause, because, right before the anomaly occurrence time, multiple actors had interacted with the same set of resources (e.g., multiple users running the same updating stack), causing confusion for the random walk backtracking. However, \toolname{} narrows down the search space by providing a small ranked list of root causes for investigation. It is worth noting that \toolname{} is the first RCL method designed to work directly with event data as defined in Section~\ref{sec:term}. Therefore, it can capture the operations performed in the systems and precisely construct the intervention graph (Algorithm~\ref{alg:construct-intervention-graph}). This graph subsequently allows our time-aware random walk to localize the root cause automatically and precisely. Meanwhile, all existing RCL methods for cloud systems~\cite{brianlogsurvey2025, soldani2022anomaly, cheng2023ai} are developed for metrics, logs, and traces, which undermines the power of event data and leaves it under-explored.

\begin{table*}[t]
\caption{The anomaly localization performance of \toolname{} and ten baselines on five datasets (Falcon, Flask, Live, \zrhEvent{}, and \mxpEvent{}) in terms of AC@1, AC@3, and Avg@5. We report the mean and standard deviation over ten runs with different random seeds. We \textbf{bold} the best values and \underline{underline} the second-best.}
\label{tab:rq2-rca}
\resizebox{\textwidth}{!}{%
\setlength\tabcolsep{1pt}
\begin{tabular}{lccccccccccccccc}
\toprule
\multirow{2}{*}{\textbf{Method}} & \multicolumn{3}{c}{\textbf{Falcon}} & \multicolumn{3}{c}{\textbf{Flask}} & \multicolumn{3}{c}{\textbf{Live}} & \multicolumn{3}{c}{\textbf{\zrhEvent{}}} & \multicolumn{3}{c}{\textbf{\mxpEvent{}}} \\ \cmidrule(lr){2-4} \cmidrule(lr){5-7} \cmidrule(lr){8-10} \cmidrule(lr){11-13} \cmidrule(lr){14-16}
 & AC@1 & AC@3 & Avg@5 & AC@1 & AC@3 & Avg@5 & AC@1 & AC@3 & Avg@5 & AC@1 & AC@3 & Avg@5 & AC@1 & AC@3 & Avg@5 \\ \midrule
\rowcolor{linegray} Random & 0.05±0.03 & 0.22±0.07 & 0.22±0.04 & 0.04±0.04 & 0.20±0.08 & 0.19±0.06 & 0.08±0.06 & 0.24±0.05 & 0.23±0.05 & 0.05±0.08 & 0.20±0.13 & 0.20±0.11 & 0.09±0.09 & 0.24±0.10 & 0.23±0.09 \\
BARO~\cite{pham2024baro} & 0.50±0.00 & 0.83±0.00 & \underline{0.73±0.00} & \underline{0.67±0.00} & 0.83±0.00 & \underline{0.81±0.00} & 0.80±0.00 & \textbf{1.00±0.00} & \underline{0.92±0.00} & \underline{0.20±0.00} & 0.20±0.00 & 0.28±0.00 & 0.70±0.00 & \textbf{1.00±0.00} & \underline{0.90±0.00} \\ 
\rowcolor{linegray} CausalAI~\cite{salesforce_causalai23} & 0.18±0.09 & 0.40±0.10 & 0.37±0.07 & 0.14±0.05 & 0.40±0.06 & 0.40±0.04 & 0.24±0.07 & 0.55±0.07 & 0.49±0.04 & 0.00±0.00 & 0.00±0.00 & 0.00±0.00 & 0.07±0.26 & \underline{0.27±0.45} & 0.26±0.35 \\ 
CausalRCA~\cite{xin2023causalrca} & 0.00±0.00 & 0.67±0.00 & 0.44±0.00 & 0.00±0.00 & 0.00±0.00 & 0.23±0.00 & 0.00±0.00 & 0.60±0.00 & 0.44±0.00 & 0.00±0.00 & 0.00±0.00 & 0.00±0.00 & 0.00±0.00 & \textbf{1.00±0.00} & 0.60±0.00 \\ 
\rowcolor{linegray} \(\epsilon\)-Diagnosis~\cite{ediagnosis} & 0.03±0.02 & 0.12±0.06 & 0.13±0.05 & 0.03±0.03 & 0.10±0.05 & 0.08±0.04 & 0.00±0.00 & 0.00±0.00 & 0.00±0.00 & 0.04±0.05 & \underline{0.23±0.12} & 0.22±0.09 & 0.00±0.00 & 0.00±0.00 & 0.12±0.00\\ 
Groot~\cite{Wang2021Groot} & 0.33±0.00 & 0.33±0.00 & 0.33±0.00 & 0.50±0.00 & 0.50±0.00 & 0.50±0.00 & 0.40±0.00 & 0.40±0.00 & 0.40±0.00 & 0.00±0.00 & 0.00±0.00 & 0.00±0.00 & 0.00±0.00 & 0.00±0.00 & 0.00±0.00 \\ 
\rowcolor{linegray} RCD~\cite{rcd} & \underline{0.57±0.00} & 0.69±0.04 & 0.65±0.03 & 0.41±0.03 & 0.66±0.05 & 0.60±0.04 & 0.00±0.00 & 0.09±0.02 & 0.11±0.01 & 0.02±0.04 & 0.07±0.09 & 0.05±0.07 & 0.00±0.00 & 0.20±0.40 & 0.14±0.29 \\ 
DeepHunt~\cite{sun2025interpretable} & 0.53±0.10 & 0.83±0.00 & 0.65±0.06 & 0.53±0.12 & \underline{0.89±0.07} & 0.73±0.05 & \underline{0.82±0.09} & \textbf{1.00±0.00} & 0.90±0.05 & 0.00±0.00 & 0.00±0.00 & 0.00±0.00 & 0.00±0.00 & 0.00±0.00 & 0.04±0.00 \\
\rowcolor{linegray} TVDiag~\cite{xie2025tvdiag} & 0.00±0.00 & 0.67±0.00 & 0.40±0.00 & 0.00±0.00 & 0.00±0.00 & 0.00±0.00 & 0.00±0.00 & 0.00±0.00 & 0.00±0.00 & - & - & - & - & - & -\\ 
KPIRoot~\cite{gu2024kpiroot} & 0.43±0.00 & \underline{0.90±0.00} & 0.63±0.00 & 0.13±0.00 & 0.67±0.00 & 0.41±0.00 & 0.60±0.00 & \underline{0.88±0.00} & 0.74±0.00 & 0.00±0.00 & \textbf{1.00±0.00} & \underline{0.50±0.00} & \underline{0.80±0.00} & \textbf{1.00±0.00} & 0.87±0.00\\
\midrule
\rowcolor{lineblue} \textbf{\toolname{}} & \textbf{0.70±0.02} & \textbf{1.00±0.00} & \textbf{0.91±0.00} & \textbf{0.68±0.01} & \textbf{1.00±0.00} & \textbf{0.91±0.01} & \textbf{1.00±0.00} & \textbf{1.00±0.00} & \textbf{1.00±0.00} & \textbf{1.00±0.00} & \textbf{1.00±0.00} & \textbf{1.00±0.00} & \textbf{1.00±0.00} & \textbf{1.00±0.00} & \textbf{1.00±0.00} \\ \bottomrule
\end{tabular}
}

{\tiny (-) TVDiag~\cite{xie2025tvdiag} requires failure training samples. The OUT and AVA incidents are test data without corresponding failure training samples.}
\vspace{-10pt}
\end{table*}

\textbf{(2) Metric-based RCL methods show reasonable performance but fail to fully exploit event data.} BARO and $\epsilon$-Diagnosis are metric-based RCL methods. To apply them, we transform event data into time series reflecting the frequency of actor interventions, where the number of series depends on the number of actors active during the incident. BARO consistently achieves the second-best performance. Its core assumption is that the root cause exhibits a strong anomaly during the incident (i.e., the most anomalous time series is considered the root cause). This assumption is surprisingly effective, but does not generalize as BARO performs poorly on the \zrhEvent{} dataset, where multiple rare interventions occurred concurrently, confusing the algorithm as BARO does not correlate interventions with anomalies as in \toolname{}. Moreover, BARO provides little interpretability, offering no explanation for why a candidate is flagged as the root cause, thereby shifting the burden of reasoning to operators. In contrast, \toolname{} does not rely on such assumptions. Instead, it constructs an intervention graph that explicitly encodes the causal relationships between recent interventions and detected anomalies, enabling precise and interpretable RCL.

\textbf{(3) Causal inference methods fail to capture semantic relationships in event data.} CausalAI, CausalRCA, RCD, and KPIRoot are causal inference--based RCL techniques, originally designed for metric time series. CausalAI, CausalRCA, and RCD apply causal discovery algorithms (e.g., RCD with $\Psi$-PC, CausalRCA with DAG-GNN) on the time series data to construct causal graphs under the assumption that the root-cause time series influences many others. Root cause ranking is then performed via graph centrality algorithms such as PageRank. KPIRoot~\cite{gu2024kpiroot} extends this paradigm by integrating both similarity analysis using Jaccard coefficient and causality analysis using Granger causality to identify sequential dependencies between KPI metrics. When adapted to event data, we transform events into frequency-based time series. KPIRoot shows inconsistent results as it achieves AC@3 scores between 0.67 and 0.90 on benchmark datasets but zero AC@1 on the \zrhEvent{} dataset. Similarly, Groot~\cite{Wang2021Groot} constructs an event causality graph and applies PageRank to identify high-impact nodes. However, Groot achieves limited accuracy with AC@3 ranging from 0.33 to 0.50 on benchmark datasets and 0.00 on real-world incidents, as it relies on predefined monitoring events rather than directly modeling actor interventions.

Despite the sophistication of these methods, they all share a fundamental limitation as they operate exclusively on time series representations, which fail to capture the semantic relationships between actors, operations, and resources encoded in events. For example, Granger causality identifies temporal precedence between time series but cannot reason about intervention logic: \textit{which} actor performed \textit{what} operation on \textit{which} resource. By contrast, the Intervention Graph in our \toolname{} leverages event information directly, explicitly encodes the relationships presented in event data that enable precise RCL through time-aware random walk.

\textbf{(4) Existing graph-based RCA approaches for microservices fail due to the mismatch between microservice dependency graphs and event-based intervention graphs.} DeepHunt~\cite{sun2025interpretable} and TVDiag~\cite{xie2025tvdiag} represent recent advances in microservice failure diagnosis. DeepHunt uses graph autoencoders (GAE) trained in a self-supervised manner on normal system behavior to learn patterns from metrics, logs, and traces. It computes root cause scores by integrating reconstruction errors with failure propagation patterns in service dependency graphs. TVDiag employs task-oriented learning with contrastive cross-modal associations to extract view-invariant failure information, but requires labeled historical failure samples from the target system for supervised training. This training data requirement becomes a critical limitation as the \zrhEvent{} and \mxpEvent{} real-world incidents serve as test data without corresponding failure training samples, preventing TVDiag from being evaluated. DeepHunt can be applied to these incidents by training on normal data from the same systems. Furthermore, TVDiag's architecture assumes a fixed system topology, making transfer learning from benchmark systems to real-world incidents infeasible.

Both methods face a fundamental mismatch when applied to event-based ADL. They are designed to operate on \textit{service dependency graphs}, constructed from traces. In contrast, the \textit{intervention graphs} constructed from events model actor--operation--resource relationships, fundamentally different semantics from service dependencies. This semantic mismatch causes severe generalization failures. DeepHunt achieves an AC@3 score of 0.83 on the Falcon dataset but drops to 0.00 AC@3 on both \zrhEvent{} and \mxpEvent{} incidents. Although self-supervised learning allows DeepHunt to avoid labeled failure data, it still requires normal training data from each target system. TVDiag compounds this limitation by requiring labeled failure samples from the target system, therefore, it cannot be evaluated on \zrhEvent{} and \mxpEvent{} incidents, which lack such training data. In contrast, \toolname{}'s unsupervised, intervention-based approach requires no training data and works across all system layers by directly modeling the event semantics, achieving 100\% AC@1 on both real-world incidents.

\vspace{-5pt}
\subsubsection{Failure Case Analysis}
We analyze failure cases where the root cause was not ranked first. The dominant failure pattern occurs when background services (e.g., logging or monitoring agents) interact with many resources within the observation window. These services create numerous edges in the Intervention Graph, forming high-connectivity nodes that accumulate more random walk visits than the root cause. In these cases, the ground truth actor typically interacts with a few resources, resulting in fewer paths for the random walker. Despite this
limitation, \toolname{} effectively narrows the search space from potentially hundreds of actors to just three candidates. Combined with the interpretable Intervention Graph, operators can quickly recognize the root cause by examining its sequence of operations and their relationships to the detected anomalies.

\begin{wraptable}{r}{0.54\textwidth}
\vspace{-30pt}
\caption{Runtime comparison of \toolname{} and baselines.} 
\vspace{-10pt}
\label{tab:rq3-runtime-split}
\centering
{\footnotesize \textbf{(a) Anomaly Detection Runtime (in seconds).}} \\
\vspace{2pt}
\resizebox{\linewidth}{!}{%
\setlength{\tabcolsep}{1pt}
\begin{tabular}{lccccc}
\toprule
\textbf{Method} & \textbf{Falcon} & \textbf{Flask} & \textbf{Live} & \textbf{\zrhEvent{}} & \textbf{\mxpEvent{}} \\
\midrule
APE~\cite{event_ijcai16_ape} & 0.060±0.08 & 0.075±0.08 & 0.141±0.14 & 0.093±0.02 & 1.290±0.19\\
\rowcolor{linegray} BARO~\cite{pham2024baro} & 0.886±0.08 & 0.913±0.06 & 0.890±0.06 & 0.883±0.06 & 0.830±0.06 \\
CUSUM~\cite{cusum} & 0.011±0.04 & 0.010±0.04 & 0.003±0.00 & 0.010±0.04 & 0.016±0.05 \\
\rowcolor{linegray} DeepSVDD~\cite{deepsvdd} & 7.780±3.59 & 10.25±7.01 & 9.500±7.74 & 5.560±2.12 & 5.416±1.75\\ 
DIF~\cite{dif} & 46.22±46.5 & 95.03±130 & 86.54±117 & 12.68±1.39 & 17.31±1.38 \\ 
\rowcolor{linegray} ICL~\cite{icl} & 110.2±144 & 245.9±404 & 229.3±329 & 27.34±3.56 & - \\ 
NeuralLog~\cite{le2021neurallog} & 4.642±2.83 & 1.375±0.11 & 0.971±0.27 & 0.338±0.02 & 3.820±0.21\\ 
\rowcolor{linegray} NeuTraL~\cite{neutral} & 11.32±6.88 & 16.79±15.6 & 18.09±20.3 & 9.290±0.56 & - \\ 
NSigma~\cite{circa} & 0.003±0.00 & 0.003±0.00 & 0.003±0.00 & 0.012±0.01 & 0.006±0.01 \\
\rowcolor{linegray} RCA~\cite{rca} & 18.91±15.1 & 32.75±39.6 & 31.25±39.2 & 12.38±5.30 & 143.0±1.91 \\ 
RDP~\cite{rdp} & 8.440±3.94 & 11.37±8.33 & 10.64±8.99 & 9.860±5.17 & - \\ 
\rowcolor{linegray} ShadeWat~\cite{zengy2022shadewatcher} & 0.147±0.75 & 0.131±0.71 & 0.047±0.22 & 0.010±0.01 & 0.570±0.18 \\ 
ADAMAS~\cite{gu2025adamas} & 4.847±0.98 & 5.547±0.99 & 4.599±0.70 & 2.839±0.26 & 92.89±1.52 \\
\rowcolor{linegray} KPIRoot~\cite{gu2024kpiroot} & 0.040±0.01 & 0.022±0.01 & 0.041±0.02 & 0.030±0.01 & 0.353±0.01\\
\midrule
\rowcolor{lineblue} \textbf{\toolname{}}  & \textbf{0.031±0.01} & \textbf{0.104±0.01} & \textbf{0.017±0.01} & \textbf{0.096±0.03} & \textbf{0.022±0.01} \\
\textit{ESP-only} & 0.001±0.00 & 0.004±0.00 & 0.007±0.00 & 0.006±0.02 & 0.006±0.02 \\ 
\textit{EFP-only} & 0.030±0.01 & 0.100±0.01 & 0.010±0.01 & 0.090±0.02 & 0.016±0.01 \\ 
\bottomrule
\end{tabular}
}

\vspace{6pt}

{\footnotesize \textbf{(b) RCL Runtime (in seconds).}} \\
\vspace{2pt}
\resizebox{\linewidth}{!}{%
\setlength{\tabcolsep}{1pt}
\begin{tabular}{lccccc}
\toprule
\textbf{Method} & \textbf{Falcon} & \textbf{Flask} & \textbf{Live} & \textbf{\zrhEvent{}} & \textbf{\mxpEvent{}} \\
\midrule

BARO~\cite{pham2024baro} & 0.120±0.01 & 0.135±0.01 & 0.290±0.03 & 0.009±0.00 & 0.010±0.00 \\
\rowcolor{linegray} \(\epsilon\)-Diagnosis~\cite{ediagnosis} & 2.320±3.25 & 2.470±3.15 & 1.620±2.02 & 1.130±0.12 & 0.160±0.05  \\
Groot~\cite{Wang2021Groot} & 9.797±0.11 & 13.81±0.22 & 8.236±0.03 & 0.357±0.01 & 0.760±0.09 \\
\rowcolor{linegray} RCD~\cite{rcd} & 15.78±36.1 & 10.79±24.1 & 9.530±17.9 & 0.190±0.02 & 81.21±0.67 \\ 
 CausalAI~\cite{salesforce_causalai23} & 81.06±52.7 & 63.35±34.9 & 73.73±44.5 & 2.580±0.01 & 1.460±0.98 \\
\rowcolor{linegray} CausalRCA~\cite{xin2023causalrca} & 185.7±296 & 189.8±299 & 232.1±341 & 55.95±4.67 & 54.66±4.47\\
DeepHunt~\cite{sun2025interpretable} & 30.10±0.31 & 33.88±0.48 & 14.24±0.31 & 4.100±0.14 & 243.0±1.90 \\
\rowcolor{linegray} TVDiag~\cite{xie2025tvdiag} & 66.90±0.22 & 73.10±0.00 & 85.11±0.00 & - & -\\
KPIRoot~\cite{gu2024kpiroot} & 554.9±0.56 & 907.4±27.2 & 445.3±13.5 & 11.37±0.29 & 600.21±0.84 \\
\midrule
\rowcolor{lineblue} \textbf{\toolname{}} & \textbf{2.366±0.06} & \textbf{4.319±0.04} & \textbf{2.420±0.04} & \textbf{0.130±0.00} & \textbf{0.034±0.00} \\
\bottomrule
\end{tabular}
}
\vspace{-30pt}
\end{wraptable}
\subsection{RQ3: How Efficient is \toolname{}?}\label{sec:eval-efficiency}

Table~\ref{tab:rq3-runtime-split} reports the runtime performance of \toolname{} against the baselines. All methods are evaluated over 10 runs across all datasets and we report their mean and standard deviation. For fairness, we report only the inference time for deep learning methods (e.g., NeuralLog, APE), excluding their significant training overheads. 

\subsubsection{Anomaly Detection}

As shown in Table~\ref{tab:rq3-runtime-split}~(a), \textbf{\toolname{} is highly efficient}, consistently outperforming most baselines across all datasets. This efficiency can be attributed to the design of its two modules: ESP, which relies on lightweight event-matching expressions, and EFP, which uses subsequence distance comparison. NSigma is extremely efficient due to its simplicity as it only considers the deviation from mean and standard deviation of the frequency. In contrast, deep learning methods are significantly slower, even though we report only their inference time. These methods require substantial computational resources during both training and inference.

\subsubsection{Root Cause Localization}

As presented in Table~\ref{tab:rq3-runtime-split}~(b), \textbf{\toolname{} can localize the root cause of anomalies within seconds.} It takes 2.4, 4.3, and 2.4 seconds to localize the root causes in the Falcon, Flask, and Live datasets, respectively. We observe that the runtime is split fairly evenly between the construction of the intervention graph and the random walk (N=100). We observe that BARO is the fastest RCL method as it is a simple statistical approach that only considers three values (median, IQR, and maximum value) across all time series, making it very efficient. On the other hand, causal inference-based methods (CausalAI, CausalRCA, RCD) are time-consuming. These methods construct causal graphs between time series, which involves computationally expensive operations such as calculating correlations between all possible pairs of time series.

\begin{wrapfigure}{r}{0.45\textwidth}
\vspace{-12pt}
\centering
\includegraphics[width=\linewidth]{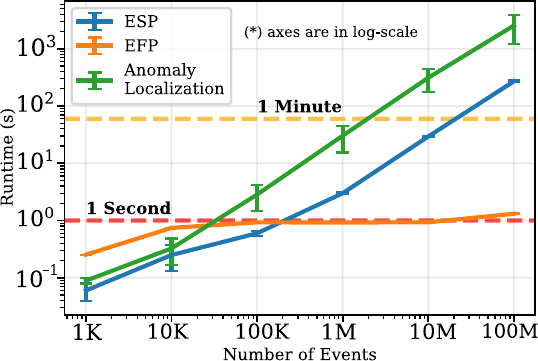}
\vspace{-15pt}
\caption{Scalability of \toolname{}.}
\label{fig:scalability}
\vspace{-10pt}
\end{wrapfigure}
\subsubsection{Scalability of \toolname{}}

In this section, we evaluate the scalability of three components in \toolname{} when handling large event streams. We deploy \toolname{} on a machine with 12 vCPUs and 36GB RAM, and measure its runtime when processing event streams from deployed systems with different scales. Figure~\ref{fig:scalability} presents the runtimes of \toolname{} across different scales. 

We observe that \toolname{} can support real-time monitoring. Specifically, ESPs and EFPs can detect anomalies at rates of 100K events/s. Recall from our real-world analysis (Figure~\ref{fig:real-world-insights}) that the number of events per incident has a median of 1K and a \(3^{rd}\) quartile of 100K. After detection, \toolname{} takes less than 1 minute to localize the root cause with up to 1M events. In addition, ESP and RCL exhibit linear growth with the number of events. This is expected, as ESP must scan the entire stream for pointwise anomaly detection and the construction of the Intervention Graph for RCL also require similar scanning. Nevertheless, both remain efficient at scale. Notably, EFP runtime remains nearly constant. Its runtime ranges from 0.25s at 1K events to only 1.3s at 100M events. The EFP module scales very favorably because EFP operates on event-based time series rather than raw events, and we observe that the number of unique time series extracted does not grow proportionally with the number of events.

\subsection{RQ4: Ablation Study} \label{sec:ablation}

We conduct ablation experiments to better understand the contribution of each component in \toolname{}. First, we examine how each component contributes to the overall anomaly detection performance, then we present empirical evidence supporting our choices in EFP design (magnitude-based over shape-based) and ESP choices (HyGLAD vs Drain). Note that the RCL component is already minimal, with no parts to isolate or remove.

\subsubsection{Anomaly Detection} \label{sec:ablation-ad}

The results in Table~\ref{tab:rq1-anomaly-detection} show that \textit{ESP-only} and \textit{EFP-only} achieve high performance and positively contribute to the overall performance of \toolname{}. We can observe that EFPs outperform other statistical methods: like \textit{EFP-only}, NSigma and CUSUM also detect anomalies in frequencies of ESPs, yet \textit{EFP-only} has a higher F1-score on average across all five datasets. We also observe that \textit{ESP-only} and \textit{EFP-only} have high recall on most datasets. This is because incidents often manifest anomalies across multiple dimensions, both pointwise and frequency-based anomalies. While our analysis in Section~\ref{sec:analysis-real-world} shows that some incidents only manifest anomalies along a single dimension, there may exist undetected anomalies in other dimensions that remain unreported. To further assess how ESP and EFP complement each other, we randomly injected 20\% anomalies into the historical (training) period for \textit{ESP-only (C)} and \textit{EFP-only (C)}. Naturally, both miss the anomalies injected in training, resulting in recall scores of only $\approx$80\% each. However, their combination—\textit{\toolname{} (C)}—achieves 96\% recall, since \textit{EFP-only (C)} is able to recover 80\% of the anomalies missed by \textit{ESP-only (C)}. This experiment shows that combining ESP and EFP helps detect subtle anomalies that may occur in only one dimension. We further show that our adaptation mechanism effectively reduces false positives while maintaining high recall. In dynamic cloud systems, without adaptation to system evolution, ESP and EFP may continuously flag false anomalies. Notably, \textit{ESP-only} and \textit{EFP-only}, when run without adaptation, achieve precision scores of 0.72 and 0.79 on the Falcon dataset, already outperforming most baselines. Our \toolname{}, which combines ESP, EFP, and adaptation, further reduces false alarms, raising precision to 0.82. Importantly, when encountering system evolution for the first time, ESP and EFP alone cannot distinguish legitimate evolution from anomalies, and will report both. With RCL, however, \toolname{} enables operators to identify the true sources of anomalies, discard false alarms, and trigger adaptation.

\begin{wrapfigure}{r}{0.42\textwidth}
\vspace{-25pt}
\centering
\includegraphics[width=\linewidth]{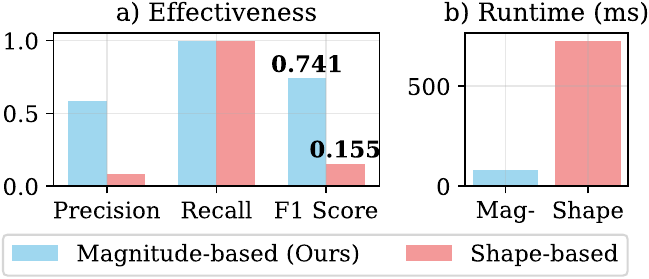}
\vspace{-17pt}
\caption{Magnitude-based vs. shape-based EFP.}
\vspace{-10pt}
\label{fig:ablation-mag}
\end{wrapfigure}
\subsubsection{Shape-based vs Magnitude-based EFP} \label{sec:ablation-shape-mag}
In this ablation, we compare our magnitude-based EFP with the shape-based variant~\cite{lu2022matrix, lee2024explainable} to assess their impact on the \zrhEvent{} dataset. As shown in Figure~\ref{fig:ablation-mag}, our magnitude-based EFP outperforms the shape-based approach with a substantially higher F1-score (0.741 vs. 0.155), demonstrating its effectiveness in detecting anomalies in event frequency (Section~\ref{sec:efp}). Moreover, it achieves a 9$\times$ speedup in runtime (74.98ms vs. 664.08ms) because our method does not require preprocessing the time series to match shapes. These results indicate that our magnitude-based subsequence comparison is better suited for event data.

\subsubsection{HyGLAD-based vs Drain-based ESPs}\label{sec:drain-study}

As discussed in Section~\ref{sec:esp}, \toolname{} can use different methods to learn ESPs. In this section, we replace HyGLAD~\cite{hyglad} with Drain~\cite{he2017drain} to examine how the performance varies. Since Drain operates on unstructured logs, we implement two variants of event-to-log conversion. In the first variant, \textit{\toolname{}(D1)}, we flatten each structured event into a log string. In the second variant, \textit{\toolname{}(D2)}, we extract key fields (time, actor, operation, resource) from the events and construct log entries in the format: \texttt{"time=<time> actor=<actor> operation=<ops> resources=<res>"}. Drain then learns the log templates, which we use as ESPs.

The experimental results presented in Table~\ref{tab:rq1-anomaly-detection} show that both variants of \toolname{} using Drain for learning ESPs still outperform most baselines consistently across all datasets, demonstrating robustness when using different methods to learn ESPs. However, as discussed in Section~\ref{sec:esp}, Drain may over-generalize some specific patterns on complex datasets (see Table~B1 in the supplementary material), as it does not take into account the relationships between system entities, resulting in missed anomalies. In the Live dataset, \toolname{}'s recall drops from 100\% to 93\% when using Drain to learn ESPs instead of HyGLAD, showing that HyGLAD is more suitable for event data.

\begin{figure*}[h]
\vspace{-5pt}
\centering
\includegraphics[width=\textwidth]{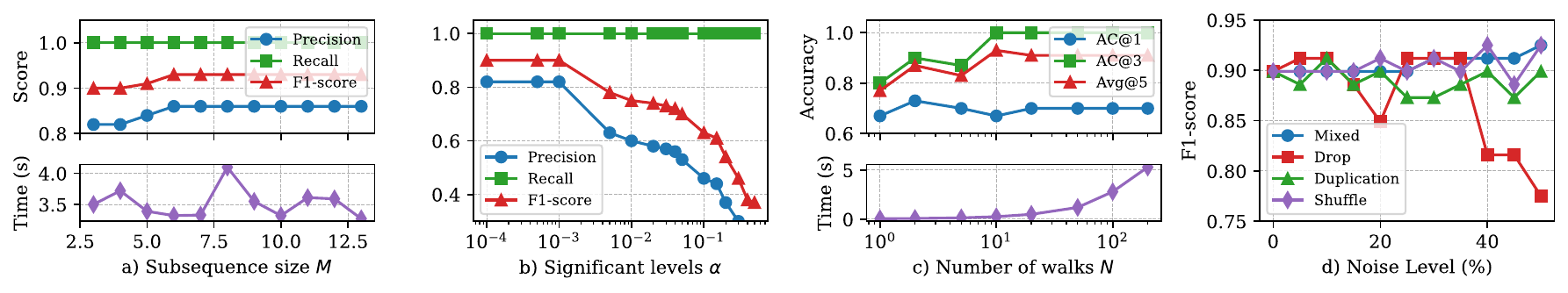}
\vspace{-20pt}
\caption{Robustness analysis of \toolname{} w.r.t. different parameters and noise levels on the Falcon dataset.}
\label{fig:sensitivity-m}
\vspace{-15pt}
\end{figure*}

\subsection{RQ5: Robustness of \toolname{}}\label{sec:robustness}

\subsubsection{Robustness to Parameter Settings}

In this section, we conduct a robustness analysis to understand how the performance of \toolname{} varies under different parameter settings. We refer readers to~\cite{hyglad} for information about parameters for ESP. EFP has two parameters: (1) the subsequence size $M$, and (2) the significance level $\alpha$. RCL has one parameter: the number of walks $N$ used in the time-aware random walk (Algorithm~\ref{alg:temporal-random-walk}).

We first vary the subsequence size $M$ from 3 to 13 to see how \toolname{} performs with both smaller and larger subsequences. Second, we vary the significance level $\alpha$ from $10^{-4}$ to 0.05 to examine the sensitivity of \toolname{}. Third, we vary the number of walks $N$ used in RCL from 1 to 200 to measure changes in performance. Due to space constraints, we only conduct these experiments on the Falcon dataset. The results are presented in Figure~\ref{fig:sensitivity-m}a, b, and c.

We observe that \toolname{} maintains stable performance across all values of $M$. However, shorter subsequences (e.g., $M < 6$) cause a slight drop in precision (from 0.86 to 0.82), as they may fail to capture more complex frequency patterns. Second, we find that \toolname{} becomes more sensitive as $\alpha$ increases (i.e., detecting more anomalies), resulting in lower precision and F1-scores. For example, precision drops from 0.82 to 0.23 as $\alpha$ increases from $10^{-4}$ to 0.05, reflecting a higher number of false positives. This finding suggests that a smaller $\alpha$ (e.g., $10^{-3}$) is preferred to avoid over-triggering alarms. Third, we observe that very small values of $N$ (e.g., 1 or 2) yield suboptimal localization performance due to limited exploration on the Intervention Graph. Accuracy improves significantly and stabilizes once $N \geq 10$, although runtime increases linearly. For instance, Avg@5 improves from 0.77 (at $N=1$) to 0.91 (at $N=20$), but runtime also increases from 0.05s to 0.5s. In practice, \toolname{} can execute RCL continuously, therefore, operators receive an initial ranked list of root causes that becomes progressively refined as additional nodes are traversed.

\subsubsection{Robustness to Noise}

We add noise to training event data by randomly (1) dropping, (2) duplicating, (3) shuffling, and (4) mixing these three, with intensity ranging from 1\% and 50\% of the total events. Figure~\ref{fig:sensitivity-m}d show that \toolname{} maintains stable performance under these noise conditions. However, it degrades under event dropping, as fewer events reduce the quality of learned patterns (e.g., F1 = 0.775 at 50\% drop). These results show that our \toolname{} is resilient to imperfect training data--a critical property for deployment in noisy, evolving cloud environments.

\vspace{-5pt}
\subsection{Generalizability of \toolname{} to Non-Event Data}
\vspace{-2pt}

\toolname{} is developed specifically for event data, where ESP and RCL are designed to leverage structured information encoded in events such as  actors, operations, and resources (Section~\ref{sec:term}). Therefore, ESP and RCL may not generalize to other data sources (e.g., metrics, logs, traces) as they do not contain this required information (see Figure~\ref{fig:event-example}). Nevertheless, EFP detects frequency-based anomalies and can be adapted to detect anomalies from other data sources. To assess the generalizability of EFP on non-event datasets, we evaluate its performance on three public benchmarks: Eadro~\cite{lee2023eadro}, GAIA~\cite{gaia}, and AIOps21~\cite{aiops21}. These datasets contain metrics, logs, and traces, which are fundamentally different from the structured event data. Eadro~\cite{lee2023eadro} provides two datasets with collected metrics, logs, and traces from two microservice systems, namely Train Ticket (Eadro TT) and Social Network (Eadro SN). Eadro TT has 27 services with 81 fault cases, and Eadro SN has 12 services with 36 fault cases.  GAIA~\cite{gaia} provides the MicroSS-Companion dataset containing 216 labeled time series with injected anomalies across 7 anomaly types (e.g., changepoint, periodic). AIOps21~\cite{aiops21} is a dataset from the AIOps Challenge 2021 with 141 fault cases across 6 fault types.

To detect frequency-based anomalies using EFP, we follow existing works~\cite{sun2024art, guo2024logformer, lee2023eadro} to extract time series from the data (metrics, logs, traces). Similar to event-based anomaly detection (Section~\ref{sec:efp}), EFP flags an anomaly if any of the time series exhibits frequency deviations during the test window. We apply EFP to all available time series in each test case. We benchmark EFP against baseline methods established in prior work on these datasets. For the Eadro datasets, we report results from the Eadro paper~\cite{lee2023eadro}, including TraceAnomaly~\cite{liu2020unsupervised}, MultimodalTrace~\cite{nedelkoski2019anomaly}, MS-RF-AD, MS-SVM-AD, MS-LSTM, and MS-DCC. For the GAIA dataset, we compare against baselines in the LogFormer paper~\cite{guo2024logformer}, including SVM, DeepLog~\cite{du2017deeplog}, LogAnomaly~\cite{meng2019loganomaly}, PLELog~\cite{yang2021semi}, LogRobust~\cite{zhang2019robust}, and LogFormer. For the AIOps21 dataset, we use baseline results from the ART paper~\cite{sun2024art}, including ART, Eadro~\cite{lee2023eadro}, and Hades~\cite{lee2023heterogeneous}. Table~\ref{tab:public-benchmarks} presents the results. We observe that:

\begin{table*}[h]
\vspace{-5pt}
\centering
\caption{Performance comparison on public benchmarks: Eadro, GAIA, and AIOps21.} \label{tab:public-benchmarks}
\vspace{-8pt}
\resizebox{\textwidth}{!}{%
\setlength\tabcolsep{3pt}
\begin{tabular}{ccccccccccccccc}
\toprule
\multirow{2}{*}{\textbf{Method}} & \multicolumn{3}{c}{\textbf{Eadro TT}} & \multicolumn{3}{c}{\textbf{Eadro SN}} & \multirow{2}{*}{\textbf{Method}} & \multicolumn{3}{c}{\textbf{GAIA}} & \multirow{2}{*}{\textbf{Method}} & \multicolumn{3}{c}{\textbf{AIOps21}} \\
\cmidrule(lr){2-4} \cmidrule(lr){5-7} \cmidrule(lr){9-11} \cmidrule(lr){13-15}
& Precision & Recall & F1-Score & Precision & Recall & F1-Score & & Precision & Recall & F1-Score & & Precision & Recall & F1-Score \\
\midrule
\rowcolor{linegray} TraceAnomaly~\cite{liu2020unsupervised} & 0.486 & 0.414 & 0.589 & 0.539 & 0.468 & 0.636 & SVM~\cite{guo2024logformer} & 0.210 & 0.540 & 0.300 & ART~\cite{sun2024art} & 0.877 & 0.960 & 0.917 \\
MultimodalTrace~\cite{nedelkoski2019anomaly} & 0.608 & 0.576 & 0.644 & 0.676 & 0.632 & 0.726 & DeepLog~\cite{du2017deeplog} & 0.180 & 0.820 & 0.310 & Eadro~\cite{lee2023eadro} & 0.767 & 0.935 & 0.842 \\
\rowcolor{linegray} MS-RF-AD~\cite{lee2023eadro} & 0.817 & 0.705 & 0.971 & 0.773 & 0.866 & 0.700 & LogAnomaly~\cite{meng2019loganomaly} & 0.230 & 0.800 & 0.360 & Hades~\cite{lee2023heterogeneous} & 0.867 & 0.868 & 0.868 \\
MS-SVM-AD~\cite{lee2023eadro} & 0.787 & 0.678 & 0.938 & 0.789 & 0.770 & 0.808 & PLELog~\cite{yang2021semi} & 0.810 & 0.860 & 0.840 & - & - & - & - \\
\rowcolor{linegray} MS-LSTM~\cite{lee2023eadro} & 0.967 & 0.997 & 0.940 & 0.948 & 0.959 & 0.937 & LogRobust~\cite{zhang2019robust} & 0.830 & 0.940 & 0.880 & - & - & - & -\\
MS-DCC~\cite{lee2023eadro} & 0.965 & 0.993 & 0.938 & 0.948 & 0.962 & 0.934 & LogFormer~\cite{guo2024logformer} & 0.890 & 0.980 & 0.930 & - & - & - & -\\
\midrule
\rowcolor{lineblue} \textbf{\toolname{}} & \textbf{0.745} & \textbf{0.975} & \textbf{0.845} & \textbf{0.857} & \textbf{1.000} & \textbf{0.923} & \textbf{\toolname{}} & \textbf{1.000} & \textbf{0.825} & \textbf{0.904} & \textbf{\toolname{}} & \textbf{0.870} & \textbf{0.930} & \textbf{0.899} \\
\bottomrule
\end{tabular}%
}
\vspace{-7pt}
\end{table*}

\textbf{(1) EFP achieves high recall on Eadro datasets for detecting frequency-based anomalies.} EFP achieves F1 of 84.5\% and 92.3\% on Eadro TT and Eadro SN, respectively, with recall scores of 97.5\% and 100\%. These datasets contain simple fault patterns as identified in~\cite{fang2025rethinkingevaluationmicroservicerca}, enabling high recall. MS-LSTM~\cite{lee2023eadro} and MS-DCC~\cite{lee2023eadro}, deep learning baselines, achieve higher F1 scores of 94.0\% and 93.8\% on Eadro TT, and 93.7\% and 93.4\% on Eadro SN. However, these methods require training data with labeled failure samples, while EFP operates in a fully unsupervised manner. 

\textbf{(2) EFP maintains competitive performance on GAIA and AIOps21.} On GAIA~\cite{gaia}, EFP achieves F1 of 90.4\% with 100\% precision. EFP performs best on magnitude-based anomalies, achieving F1 of 98.5\% on changepoint anomalies and 96.9\% on partially stationary patterns. LogFormer~\cite{guo2024logformer}, a transformer-based method, achieves 93.0\% F1. On AIOps21~\cite{aiops21}, EFP achieves F1 of 89.9\% with balanced precision (87.0\%) and recall (93.0\%), outperforming Eadro (84.2\%) and Hades (86.8\%), while ART~\cite{sun2024art} achieves 91.7\% F1. We note that some GAIA anomalies are not visible upon manual inspection~\cite{GAIAIssue11}, and a recent evaluation~\cite{fang2025rethinkingevaluationmicroservicerca} excluded GAIA due to ground-truth inconsistencies.

\section{Threats to Validity}

We assess potential threats to the validity of our work, following the guidelines outlined by Wohlin et al.~\cite{wohlin2012experimentation}. The \textbf{construct validity} primarily concerns the hyperparameter settings and evaluation metrics. To mitigate this, we use established evaluation metrics and adopt recommended configurations from previous works~\cite{chen2022adsketch, pham2024baro, pham2026torai, pham2025rcaeval}. Another threat lies in the use of ESPs, which may be susceptible to adversarial evasion (e.g., attackers may mimic normal patterns to bypass detection). However, such evasive behavior likely triggers other consequences, which will eventually be detected by our framework. The \textbf{internal validity} stems from potential implementation bugs that could affect result reliability. We mitigate this by using well-maintained Python libraries, extensive testing, and repeating each experiment multiple times to ensure consistency. The \textbf{conclusion validity} stems from our benchmark datasets not covering the full range of anomaly types. While we base our incident reproduction on a systematic analysis of 520 real-world incident reports, certain scenarios require manual intervention because they fall outside \toolname{}'s design scope. For anomaly detection, issues that do not manifest through event data cannot be detected. For example, a hardware fault (e.g., disk full) causing a node crash may not generate events. Similarly, performance degradation captured only in metrics (e.g., increased latency) without corresponding event signatures would be missed. For RCL, interventions with low connectivity in the Intervention Graph may receive fewer random walk visits than unrelated high-activity nodes, as discussed in Section~\ref{sec:experiments}. Nevertheless, \toolname{} still narrows the search space by identifying affected resources and recent interventions, enabling operators to extend their investigation. The \textbf{external validity} concerns the generalizability of our findings. In this study, we deployed our method in real cloud systems and evaluated against real incident data, grounding evaluation in realistic settings.

\section{Conclusion}

We present \toolname{}, the first open-box anomaly detection and RCL framework designed for event data in cloud systems. Our real-world incident analysis provides the empirical foundation for this work, revealing that event-based anomalies manifest through Event Type, Event Value, and Event Frequency, and their root causes require tracing intervention chains. Guided by these findings, \toolname{} detects pointwise anomalies through \textit{Event Semantic Patterns (ESPs)} and frequency-based anomalies through \textit{Event Frequency Patterns (EFPs)}, and localizes root causes by constructing an \textit{Intervention Graph} and performing a time-aware random walk. Our evaluation on three benchmark systems and two real-world incidents demonstrates that \toolname{} achieves F1-scores of at least 90\% for anomaly detection and 100\% top-3 accuracy for RCL. We further show that EFP generalizes to non-event data, achieving competitive performance on three public benchmarks. We release our event datasets to facilitate future research on event-based ADL.

\section*{Data Availability}\label{sec:data-availability}
The implementation of \toolname{}, the three experimental datasets, and the supplementary materials are available on Zenodo at \url{https://zenodo.org/records/19433493}~\cite{eventadl_artifact}.

\begin{acks}
This work was done during an internship at Amazon. Luan Pham was a PhD student at RMIT University under the Australian Research Council Discovery Project (DP220103044).
\end{acks}

\bibliographystyle{ACM-Reference-Format}
\bibliography{references}

\end{document}